\documentclass[11pt,a4paper]{article}
\usepackage{times,latexsym}
\usepackage{url}
\usepackage[table]{xcolor}
\usepackage[utf8]{inputenc}
\usepackage[T2A,LAE,T1]{fontenc}
\usepackage{graphicx}
\usepackage{tabu}
\usepackage{multirow}
\usepackage{arydshln}
\usepackage{times}
\usepackage{latexsym}
\usepackage{amssymb}
\usepackage{amsmath}
\usepackage{xspace} 
\usepackage{tablefootnote}
\usepackage{booktabs}
\usepackage{lingmacros}
\usepackage{colortbl}
\usepackage{subfig} 
\usepackage[russian,arabic,USenglish]{babel}
\usepackage{tipa}
\usepackage{color}
\usepackage{xcolor}

\expandafter\def\expandafter\UrlBreaks\expandafter{\UrlBreaks
  \do\a\do\b\do\c\do\d\do\e\do\f\do\g\do\h\do\i\do\j%
  \do\k\do\l\do\m\do\n\do\o\do\p\do\q\do\r\do\s\do\t%
  \do\u\do\v\do\w\do\x\do\y\do\z\do\A\do\B\do\C\do\D%
  \do\E\do\F\do\G\do\H\do\I\do\J\do\K\do\L\do\M\do\N%
  \do\O\do\P\do\Q\do\R\do\S\do\T\do\U\do\V\do\W\do\X%
  \do\Y\do\Z}

\usepackage{soulutf8}
\usepackage[normalem]{ulem}

\usepackage{enumitem}
\setlist{nolistsep}
\usepackage{microtype}

\iftrue 
\usepackage[final]{naacl2021}
\usepackage[final]{proofing}

  \renewcommand\hl[1]{{#1}}  
   {\draftnote{\red{#2}}}
   \newcommand\redHL[1]{}
  \newcommand\todo[1]{}
  
\newcommand{\Djame}[1]{}
\newcommand{\BM}[1]{}
\newcommand{\BS}[1]{}
\newcommand{\Antonis}[1]{}

\newcommand{\an}[1]{}
\newcommand{\bm}[1]{}

\else  

\usepackage{naacl2021}
\usepackage[draft]{proofing}

\newcommand{\Djame}[1]{
\textbf{\textcolor{red}{\hl{Djame: #1}}}
}

\newcommand{\BM}[1]{
\textbf{\textcolor{black}{\textcolor{green}{BM: #1}}}
}

\newcommand{\BS}[1]{
\textbf{\textcolor{blue}{\hl{BS: #1}}}
}

\newcommand{\Antonis}[1]{
\textbf{\textcolor{blue}{\hl{AA: #1}}}
}

\newcommand{\an}[1]{\textcolor{magenta}{[#1 --AA]}}
\newcommand{\bm}[1]{\textcolor{red}{[#1 --BM]}}

\newcommand\red[1]{{\textbf{\textcolor{red}{#1}}}}

\let\oldred\red
\renewcommand\red[1]{{\bf \oldred{{#1}}}}

 \newcommand\redHL[1]{\red{\hl{#1}}}
\let\olddraftnote\draftnote
\renewcommand\draftnote[1]{\olddraftnote{\red{#1}}}

\fi

\newcommand{\xlmr}{\mbox{XLM-R}\xspace}
\newcommand{\mbert}{\mbox{mBERT}\xspace}
\newcommand{\mbertmlm}{\mbox{mBERT+MLM}\xspace}
\newcommand{\monomlm}{\mbox{MLM}\xspace}
\newcommand{\mlmtuning}{\textsc{MLM-tuning}\xspace}
\newcommand{\tasktuning}{\textsc{Task-tuning}\xspace}

\definecolor{lightgreen}{rgb}{0.67, 0.94, 0.82}

\usepackage{xspace,mfirstuc,tabulary}

\usepackage{booktabs}
\usepackage{multirow}
\usepackage{xcolor}
\usepackage{soulutf8}

\title{When Being Unseen from \mbert is just the Beginning: \\
Handling New Languages With Multilingual Language Models}

\author{Benjamin Muller\textsuperscript{$\dagger$*} \quad Antonis Anastasopoulos\textsuperscript{$\ddagger$}
\quad Beno\^{\i}t Sagot\textsuperscript{$\dagger$} \quad Djam\'e Seddah\textsuperscript{$\dagger$} \\
  \textsuperscript{$\dagger$}Inria, Paris, France
  \quad
  \textsuperscript{*}Sorbonne Universit\'e, Paris, France\\
  \textsuperscript{$\ddagger$}Department of Computer Science, George Mason University, USA\\
  \texttt{firstname.lastname@inria.fr} \quad
  \texttt{antonis@gmu.edu}}

\date{}

\begin{document}
\maketitle
\begin{abstract}

Transfer learning based on pretraining language models on a large amount of raw data has become a new norm to reach state-of-the-art performance in NLP. Still, it remains unclear how this approach should be applied for unseen languages that are not covered by any available large-scale multilingual language model and for which only a small amount of raw data is generally available. In this work, by comparing multilingual and monolingual models, we show that such models behave in multiple ways on unseen languages. Some languages greatly benefit from transfer learning and behave similarly to closely related high resource languages whereas others apparently do not. Focusing on the latter,  we show that this failure to transfer is largely related to the impact of the script used to write such languages. We show that transliterating those languages significantly improves the potential of large-scale multilingual language models on downstream tasks. This result provides a promising direction towards making these massively multilingual models useful for a new set of unseen languages.\footnote{Code available at \url{https://github.com/benjamin-mlr/mbert-unseen-languages.git}}
\end{abstract}

\section{Introduction}

Language models are now a new standard to build state-of-the-art Natural Language Processing (NLP) systems. In the past year, monolingual language models have been released for more than 20 languages including Arabic, French, German, and Italian
~\cite[inter alia]{antoun2020arabert,martin-etal-2020-camembert,de2019bertje,canete-2020-beto,kuratov-2019-rubert,schweter-2020-berturk}. Additionally, large-scale multilingual models covering more than 100 languages are now available (\xlmr by \citet{conneau-etal-2020-unsupervised} and \mbert by \citet{devlin-etal-2019-bert}). 
Still, most of the 6500+ spoken languages in the world~\cite{hammarstrom2016linguistic}
are not covered---remaining unseen---by those models. Even languages with millions of native speakers like Sorani Kurdish (about 7 million speakers in the Middle East) or Bambara (spoken by around 5 million people in Mali and neighboring countries) are not covered by any available language models at the time of writing. 

Even if training multilingual models that cover more languages and language varieties is tempting, the curse of multilinguality \citep{conneau-etal-2020-unsupervised} makes it an impractical solution, as it would require to train ever larger models. Furthermore, as shown by \citet{wu-dredze-2020-languages}, large-scale multilingual language models are sub-optimal\draftremove{performance} for languages that \draftreplace{only account for a small portion of the pretraining.}{are under-sampled during pretraining.} 

In this paper, we analyze task and language adaptation experiments to get usable language model-based representations for under-studied low resource languages. 
We run experiments on 15 typologically diverse \draftremove{unseen}languages on three NLP tasks: part-of-speech (POS) tagging, dependency parsing (DEP) and named-entity recognition (NER).

Our results bring forth a diverse set of behaviors that we classify in three categories reflecting the abilities of pretrained multilingual language models to be used for low-resource languages\draftreplace{.}{. We dub those categories Easy, Intermediate and Hard.}\draftremove{Some languages, the ``Easy'' ones
}

Hard languages include both stable and endangered languages, but they predominantly are languages of communities that are majorly under-served by modern NLP. Hence, we direct our attention to these Hard languages.
For those languages, we show that the script they are written in can be a critical element in the transfer abilities of pretrained multilingual language models. \draftnote{Not really. .. -BS
} 
Transliterating them leads to large gains in performance \draftremove{leading to }outperforming non-contextual strong baselines. 
To sum up, our contributions are the following:

\begin{itemize}
    \item \draftremove{Based on our empirical results, }We propose a new categorization of the low-resource languages that are unseen by available language models: the Hard, the Intermediate and the Easy languages. 
    \item We show that Hard languages can be better addressed by transliterating them 
    into a better-handled script (typically Latin)
    , providing a promising direction towards making multilingual language models useful for a new set of unseen languages. 
\end{itemize}


\section{Background and Motivation}
 
As \citet{joshi2020state} vividly illustrate, there is a large divergence in the coverage of languages by NLP technologies. The majority of the 6500+ of the world's languages are not studied by the NLP community, since most have few or no annotated datasets, making systems' development challenging.

The development of such models is a matter of high importance for the inclusion of communities, the preservation of endangered languages 
and more generally to support the rise of tailored NLP ecosystems for such languages
\cite{schmidt-wiegand-2017-survey,myanmar2018,seddah-etal-2020-building}.
In that regard, the advent of the Universal Dependencies project \citep{nivre2016universal} and the WikiAnn dataset \citep{pan-etal-2017-cross} have greatly increased the number of covered languages by providing annotated datasets for more than 90 languages for dependency parsing and 282 languages for NER. 

Regarding modeling approaches, the emergence of multilingual representation models, first with static word embeddings \citep{ammar2016massively} and then with language model-based contextual representations \cite{devlin-etal-2019-bert, conneau-etal-2020-unsupervised} enabled transfer from high to low-resource languages, leading to significant improvements in downstream task performance \citep{rahimi-etal-2019-massively,kondratyuk-straka-2019-75}\draftnote{\bm{add more  citations}}. 
Furthermore, in their most recent forms, these multilingual models \draftremove{, such as \mbert,} process tokens at the sub-word level~\draftremove{using for instance SentencePiece tokenization} \citep{kudo-richardson-2018-sentencepiece}. 
As such, they work in an open vocabulary setting, only constrained by the pretraining character set. 
This flexibility enables such models to process any language, even those that are not part of their pretraining data.

When it comes to low-resource languages, one direction is to simply train contextualized embedding models on whatever data is available. Another option is to adapt/fine-tune a multilingual pretrained model to the language of interest. We briefly discuss these two options.

\paragraph{Pretraining language models on a small amount of raw data}
Even though the amount of pretraining data seems to correlate with downstream task performance (e.g. compare BERT and RoBERTa \citep{liu2020roberta})\draftnote{I don't understand this part. Seems like smthg is missing. Refs ? work that does the comparison? -ds \bm{one of the contribution of Roberta was that more data is better: I added Roberta Ref}}, several attempts have shown that training a model from scratch  can be efficient even if the amount of data in that language is limited. Indeed, \citet{suarez2020monolingual} showed that pretraining ELMo models \citep{Peters:2018} on less than 1GB of text leads to state-of-the-art performance while \citet{martin-etal-2020-camembert} showed\draftremove{ for French} that pretraining a BERT model on as few as 4GB of diverse enough data results in state-of-the-art performance. \citet{Micheli_et_al:2020:smallbertmodels} further demonstrated that decent performance was achievable with only 100MB of raw text data. 

\paragraph{Adapting large-scale models for low-resource languages} Multilingual language models can be used directly on unseen languages, or they can also be adapted using unsupervised methods. For example, \citet{han-eisenstein-2019-unsupervised} successfully used unsupervised model adaptation of the English BERT model to Early Modern English for sequence labeling.\draftnote{those 3 sentences don't read well. telegraphic style -ds}
Instead of fine-tuning the whole model, \citet{pfeiffer-etal-2020-mad} recently showed that adapter layers \citep{houlsby2019parameter} can be injected into multilingual language models to provide parameter efficient task and language transfer.


Still, as of today, the availability of monolingual or multilingual language models is limited to approximately 120 languages, leaving many languages without access to valuable NLP technology, although some are spoken by  millions of people, including Bambara and Sorani Kurdish, or are an official language of the European Union, like Maltese.

\paragraph{What can be done for unseen languages?} Unseen languages strongly vary in the amount of available data, in their script (many languages use non-Latin scripts such as Sorani Kurdish and Mingrelian), and in their morphological or syntactical properties (most largely differ from high-resource Indo-European languages). This makes the design of a methodology to build contextualized models for such languages \draftreplace{very challenging}{challenging at best}. 
In this work, by experimenting with 15 typologically diverse unseen languages, (i)~we show that there is a diversity of behavior depending on the script, the amount of available data, and the relation to the pretraining languages; (ii)~Focusing on the unseen languages that lag in performance compared to their easier-to-handle counterparts, we show that the script plays a critical role in the transfer abilities of multilingual language models. Transliterating such languages to a script which is used by a related language seen during pretraining \draftnote{This is not the case for Sorani Kurdish...
-BS fixed} can lead to  significant improvement in downstream performance.

\section{Experimental Setting}

We will refer to any languages that are not covered by pretrained 
language models as ``unseen.''
We select a small portion of those languages within a large scope of language families and scripts. Our selection is constrained to 15 typologically diverse languages for which we have evaluation data for at least one of our three downstream tasks. Our selection includes low-resource Indo-European and Uralic languages, as well as members of the Bantu, Semitic, and Turkic families. None of these 15 languages are included in the pretraining corpora of \mbert. 
Information about their scripts, language families, and amount of available raw data can be found in the Appendix in Table~\ref{tab:langs}.


\subsection{Raw Data}

To perform pretraining and fine-tuning on monolingual data, we use the deduplicated datasets from the \mbox{OSCAR} project~\cite{OrtizSuarezSagotRomary2019}. \mbox{OSCAR} is a corpus extracted from a Common Crawl Web snapshot.\footnote{\url{http://commoncrawl.org/}} 
It provides a significant amount of data for all the unseen languages we work with, except for Buryat, Meadow Mari, Erzya and Livvi for which we use Wikipedia dumps and for Narabizi, Naija and Faroese, for which we use data collected by \citet{seddah-etal-2020-building}, \citet{caron2019surface} and \citet{biemann2007leipzig} respectively.

\subsection{Non-contextual Baselines}
For parsing and POS tagging, we use the UDPipe future system~\citep{straka2018udpipe} as our baseline. This model is a LSTM-based \citep{hochreiter1997long} recurrent architecture trained with pretrained static word embedding \citep{mikolov2013distributed} (hence our non-contextual characterization) along with character-level embeddings. This system was ranked in the very first positions for parsing and tagging in the CoNLL shared task 2018 \citep{conll-2018-conll}. For NER we use the LSTM-CRF model with character and word level embedding using \citet{qi2020stanza} implementation.

\subsection{Language Models}
\label{subsec:LMs}

In all our study, we train our language models using the Transformers library 
\citep{wolf-etal-2020-transformers}. 
\paragraph{\monomlm from scratch}

The first approach we evaluate is to train a dedicated language model from scratch on the available raw data we have. To do so, we train a language-specific SentencePiece tokenizer \citep{kudo-richardson-2018-sentencepiece} before training a Masked-Language Model (\monomlm) using the RoBERTa (base) architecture and objective functions \cite{liu2019roberta}. As we work with significantly smaller pretraining sets than in the original setting, we reduce the number of layers to 6 layers in place of the original 12 layers. 

\paragraph{Multilingual Language Models}
We want to assess how large-scale multilingual language models can be used and adapted to languages that are not in their pretraining corpora. We work with the multilingual version of BERT (\mbert) trained on the concatenation of Wikipedia corpora in 104 languages \citep{devlin-etal-2019-bert}. We also ran experiments with the \xlmr base version \citep{conneau-etal-2020-unsupervised} trained on 100 languages using data extracted from the Web. As the observed behaviors\draftnote{\bm{should we talk about Uyghur which is supported by XLMR and deliver sota on uyghur (upper-bound for our experiments)}} are very similar between both models, we only report results using \mbert. Note that \mbert is highly biased toward Indo-Europeans languages written in the Latin script. More than 77\% of the subword vocabulary are in the Latin script while only 1\% are in the Georgian script~\cite{acs2019exploring}.\draftremove{The basic statistics of the vocabulary shows that more than 77\% of the vocabulary subword types are in the Latin script, about 11.5\% are in the Cyrillic script, the Arabic scripts takes up about 4\%, and smaller scripts like the Georgian one make up less than 1\% of the vocabulary (with less than 1,000 subwords)}

\paragraph{Adapting Multilingual Language Models to unseen languages with \mlmtuning}
Following previous work~\cite{han-eisenstein-2019-unsupervised,karthikeyan2019cross,pfeiffer-etal-2020-mad}, we adapt large-scale multilingual models by fine-tuning them with their Mask-Language-Model objective directly on the available raw data in the unseen target language. We refer to this process as \mlmtuning.  We will refer to a MLM-tuned \mbert model as \mbertmlm.

\begin{table*}[ht!]
\centering
\resizebox{\linewidth}{!}{
\begin{tabu}{ l  c  c c c@{\hspace{0.35cm}}  @{\hspace{0.35cm}} c  c c c@{\hspace{0.35cm}}  @{\hspace{0.35cm}} c  c c c}
	\toprule
	& \multicolumn{4}{c @{\hspace{0.5cm}}}{\textsc{UPOS}} & \multicolumn{4}{c @{\hspace{0.7cm}}}{\textsc{LAS}} & \multicolumn{4}{c @{\hspace{0.7cm}}}{\textsc{NER}} \\ 
	\cmidrule(l{2pt}r{0.4cm}){2-5}\cmidrule(l{-0.2cm}r{0.4cm}){6-9}\cmidrule(l{-0.2cm}r{0.4cm}){10-13}
	\multirow{-2}{*}[1pt]{Model} & 
	\textsc{\mbert}     &\textsc{\mbertmlm} & \monomlm & Baseline 
	& \textsc{\mbert}   &\textsc{\mbertmlm} & \monomlm & Baseline 
	& \textsc{\mbert}     &\textsc{\mbertmlm} & \monomlm & Baseline 
	\\
	\midrule
	Faroese  & 96.3 & 96.5     & 91.1& 95.4 & 84.0 & 86.4 & 67.6 & 83.1 & 52.1   &  58.3 & 39.3 & 44.8 \\
	Naija & 
	89.3 & 89.6 & 87.1 & 89.2 &
	71.5 & 69.2 & 63.0 & 68.3 &
	- & - & - & -  \\
	Swiss German  & 76.7 & 78.7  & 65.4 & 75.2 &  41.2 & 69.6 & 30.0  & 32.2 & - & - & - & - \\
	Mingrelian & - & - & - & - & - & - & - & - & 53.6 & 68.4 & 42.0 & 48.2 \\

	\bottomrule
	
\end{tabu}
}
\caption{\textbf{Easy Languages} POS, Parsing and NER scores comparing \mbert, \mbertmlm  and monolingual \monomlm to strong non-contextual baselines when trained and evaluated on unseen languages. Easy Languages are the ones on which \mbert outperforms strong baselines out-of-the-box. 
Baselines are LSTM based models from UDPipe-future~\citep{straka2018udpipe} for parsing and POS tagging and Stanza~\citep{qi2020stanza} for NER.
%
}%
\label{tab:easy}
\end{table*}

\subsection{Downstream Tasks}

We perform experiments on POS tagging, Dependency Parsing (DEP), and Name Entity Recognition (NER). We use annotated data from the Universal Dependency project \citep{nivre2016universal} for POS tagging and parsing, and the WikiAnn dataset \citep{pan-etal-2017-cross} for NER.
For POS tagging and NER, we append a linear classifier layer on top of the language model. For parsing, following \citet{kondratyuk-straka-2019-75}, we append a Bi-Affine Graph prediction layer \citep{dozat2016deep}.  We refer to the process of fine-tuning a language model in a task-specific way as \tasktuning.\footnote{Details about optimization can be found in Appendix~\ref{sec:reproducibility}}

\subsection{Dataset Splits}
For each task and language, we use the provided training, validation and test dataset split except for the ones that have less than 500 training sentences. In this case, we concatenate the training and test set and perform 8-folds cross-Validation and use the validation set for early stopping. If no validation set is available, we isolate one of the folds for validation and report the test scores as the average of the other folds. This \draftreplace{uses to run training}{enables us to train} on at least 500 sentences in all our experiments (except for Swiss German for which we only have 100 training examples) and reduce the impact of the annotated dataset size on our analysis. Since cross-validation results in training on very limited number of examples, we refer to training in this cross-validation setting as \textit{few-shot} 
learning.

\begin{figure}[h]
\includegraphics[width=8cm]{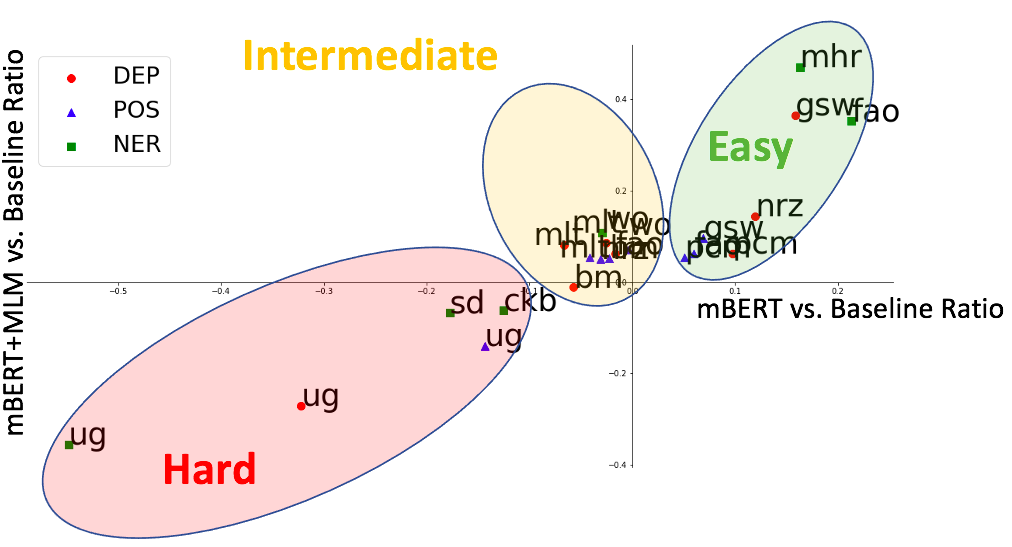}
\centering
\caption{Visualizing our Typology of Unseen Languages. (X,Y) positions are computed for each language and each task as follows: given the score of \mbert denoted $s$ and $s_0$ the baseline score: $X$~=~$\frac{s_{mBERT}-s_{0}}{s_{0}}$, $Y$~=~$\frac{s_{mBERT+MLM}-s_{0}}{s_{0}}$\\
Easy Languages are the ones on which \mbert performs better than the baseline without \mlmtuning and Intermediate languages are the ones that require \mlmtuning. For Hard languages, \mbert under-performs the baselines in all settings.}
\label{fig:typology_sumup}
\end{figure}

\section{The Three Categories of Unseen Languages}

For each unseen language and each task, we experiment with our three modeling approaches: (a)~{\bf Training a language model from scratch on the available raw data} and then fine-tuning it on any available annotated data in the target language. (b)~{\bf Fine-tuning \mbert with \tasktuning} directly on the target language. (c)~Finally, {\bf adapting \mbert to the unseen language using \mlmtuning} before fine-tuning it in a supervised way on the target language. We then compare all these experiments to our non-contextual strong baselines. By doing so, we can assess if language models are a practical solution to handle each of these unseen languages. 

Interestingly we find a large diversity of behaviors across languages regarding those language model training techniques. As summarized in Figure
 \ref{fig:typology_sumup}, we observe three clear clusters of languages.

\begin{table*}[ht!]
\centering\small
\resizebox{\linewidth}{!}{
\begin{tabu}{ l  c  c c c@{\hspace{0.35cm}}  @{\hspace{0.35cm}} c  c c c@{\hspace{0.35cm}}  @{\hspace{0.35cm}} c  c c c}
	\toprule
	& \multicolumn{4}{c @{\hspace{0.5cm}}}{\textsc{UPOS}} & \multicolumn{4}{c @{\hspace{0.7cm}}}{\textsc{LAS}} & \multicolumn{4}{c @{\hspace{0.7cm}}}{\textsc{NER}} \\ 
	\cmidrule(l{2pt}r{0.4cm}){2-5}\cmidrule(l{-0.2cm}r{0.4cm}){6-9}\cmidrule(l{-0.2cm}r{0.4cm}){10-13}
	\multirow{-2}{*}[1pt]{Model} & 
	\textsc{\mbert}     &\textsc{\mbertmlm} & \monomlm & Baseline 
	& \textsc{\mbert}   &\textsc{\mbertmlm} & \monomlm & Baseline 
	& \textsc{\mbert}     &\textsc{\mbertmlm} & \monomlm & Baseline 
	\\
	\midrule
	Maltese & 
	92.0 & \bf 96.4 & 92.05 &  96.0 & 
	74.4 & \bf 82.1 & 66.5 & 79.7 & 
	61.2 & \bf 66.7 & 62.5 & 63.1 \\
	

	Narabizi &
	  81.6  & \bf 84.2  &  71.3  & \bf  84.2 &
	  56.5  & \bf 57.8  &  41.8  &  52.8 &
	  -  & -  &  -  &  -  \\
	Bambara &  90.2 & \bf 92.6 & 78.1 & 92.3 &  71.8 &  75.4  & 46.4   & \bf 76.2 & -  & - & - \\
    Wolof &  92.8 & \bf 95.2 &88.4 &94.1 & 73.3&\bf 77.9& 62.8& 77.0 & - & - & - & -\\
    Erzya &  89.3 & \bf91.2 & 84.4 &  91.1 & 61.2 & \bf 66.6 & 47.8 & 65.1& - & - & - & - \\
    Livvi & 83.0 & \bf 85.5 & 81.1 & 84.1 &  36.3 &\bf 42.3 & 35.2 & 40.1 & - & - & - & - \\
    Mari & 
	-  & -  &  -  &  -  &
	-  & -  &  -  &  -  & 
	55.2 & \bf 57.6 & 44.0 & 56.1
	\\

	\bottomrule
	
\end{tabu}
}
\caption{\textbf{Intermediate Languages} POS, Parsing and NER scores comparing \mbert, \mbertmlm  and monolingual \monomlm to strong non-contextual baselines when trained and evaluated on unseen languages. Intermediate Languages are the ones for which \mbert requires \mlmtuning to outperform the baselines. 
%
}%
\label{tab:intermediate}
\end{table*}

The first cluster, which we dub ``Easy", corresponds to the languages that do not require extra \mlmtuning for \mbert to achieve good performance. \mbert has the modeling abilities to process such languages without relying on raw data and can outperform strong non-contextual baselines as such. In the second cluster, the ``Intermediate" languages require \mlmtuning. \mbert is not able to beat strong non-contextual baselines using only \tasktuning, but \mlmtuning enables it to do so. Finally, Hard languages are those on which \mbert fails to deliver any decent performance even after \textsc{MLM}- and \textsc{Task}- fine-tuning. \mbert simply does not have the capacity to learn and process such languages. 

We emphasize that our categorization of unseen languages is only based on the relative performance of \mbert after fine-tuning compared to strong non-contextual baseline models. We leave for future work the analysis of the absolute performance of the model on such languages (e.g. analysing the impact of the fine-tuning data set size on \mbert's  downstream performance).

In this section, we present our results in detail in each of these language clusters and provide insights into their linguistic properties. 

\subsection{Easy}
\label{sec:easy}

Easy languages are the ones on which \mbert delivers high performance out-of-the-box, compared to strong baselines.
We classify Faroese, Swiss German, Naija and Mingrelian as easy languages and report performance in Table~\ref{tab:easy}. 

We find that those languages match two conditions:
\begin{itemize}
    \item They are closely related to languages used during MLM pretraining
    \item These languages use the same script as such closely related languages.
\end{itemize}
Such languages benefit from multilingual models, as cross-lingual transfer is easy to achieve and hence quite effective.

More details about those languages can be found in Appendix~\ref{sec:easy_appendix}.

\subsection{Intermediate}
The second type of languages (which we dub ``Intermediate'') are generally harder to process for pretrained MLMs out-of-the-box. In particular, pretrained multilingual language models are typically outperformed by a non-contextual strong baselines. Still, \mlmtuning has an important impact and leads to usable state-of-the-art models. 

A good example of such an intermediate language is Maltese, a member of the Semitic language but using the Latin script. Maltese has not been seen by \mbert during pretraining. Other Semitic languages though, namely Arabic and Hebrew, have been included in the pretraining languages. \draftreplace{The results on Maltese are outlined in Table~\ref{tab:maltese}, where it is clear that the non-contextual baseline outperforms \mbert.}{As seen in Table~\ref{tab:intermediate}, the non-contextual baseline outperforms \mbert.} Additionally, a monolingual MLM trained on only 50K  sentences matches \mbert performance for both NER and POS tagging. However, the best results are reached with \mlmtuning: the proper use of monolingual data and the advantage of similarity to other pretraining languages render Maltese a {\em tackle-able} language \draftreplace{by outperforming our strong non-contextual baseline.}{as shown by the performance gain over our strong non-contextual baselines.}

\begin{table*}[ht!]
\small\centering
\resizebox{\linewidth}{!}{
\begin{tabu}{ l  c  c c c@{\hspace{0.35cm}}  @{\hspace{0.35cm}} c  c c c@{\hspace{0.35cm}}  @{\hspace{0.35cm}} c  c c c}
	\toprule
	& \multicolumn{4}{c @{\hspace{0.5cm}}}{\textsc{UPOS}} & \multicolumn{4}{c @{\hspace{0.7cm}}}{\textsc{LAS}} & \multicolumn{4}{c @{\hspace{0.7cm}}}{\textsc{NER}} \\ 
	\cmidrule(l{2pt}r{0.4cm}){2-5}\cmidrule(l{-0.2cm}r{0.4cm}){6-9}\cmidrule(l{-0.2cm}r{0.4cm}){10-13}
	\multirow{-2}{*}[1pt]{Model} & 
	\textsc{\mbert}     &\textsc{\mbertmlm} & \monomlm & Baseline 
	& \textsc{\mbert}   &\textsc{\mbertmlm} & \monomlm & Baseline 
	& \textsc{\mbert}     &\textsc{\mbertmlm} & \monomlm & Baseline 
	\\
	\midrule
	Uyghur  &  77.0 & 88.4 & 87.4 & \bf 90.0 &
	
	45.5 & 48.9 & 57.3 & \bf 67.9 & 
	
	24.3 & 34.6 & 41.4 & \bf 53.8\\
	Sindhi & 
	-  & -  &  -  &  -  &
	-  & -  &  -  &  -  &
	42.3  &  47.9  &  45.2  &  \bf 51.4  \\
	Sorani Kurdish&-  & -  &  -  &  -  &
	-  & -  &  -  &  -  & 70.4 & 75.6 & 80.6 & \bf 80.5 \\

	\bottomrule
	
\end{tabu}
}
\caption{\textbf{Hard Languages} POS, Parsing and NER scores comparing \mbert, \mbertmlm  and monolingual \monomlm to strong non-contextual baselines when trained and evaluated on unseen languages. Hard Languages are the ones for which \mbert fails to reach decent performance even after \mlmtuning.
%
}%
\label{tab:hard_summary}
\end{table*}

Our Maltese dependency parsing results are in line with those of~\citet{chau2020parsing}, who also showed that \mlmtuning leads to significant improvements. They also additionally showed that a small vocabulary transformation allowed fine-tuning to be even more effective and gain 0.8 LAS points more. We further discuss the vocabulary adaptation technique of~\citet{chau2020parsing} in section~\ref{sec:discussion}.

We consider Narabizi \citep{seddah-etal-2020-building}, an Arabic dialect spoken in North-Africa written in the Latin script and code-mixed with French, to fall in the same Intermediate category, because it follows the same pattern. \draftremove{Our results in Narabizi are listed in Table~\ref{tab:narabizi}.} 
For both POS tagging and parsing, the multilingual models outperform the monolingual NarabiziBERT. In addition, \mlmtuning leads to significant improvements over the non-language-tuned \mbert baseline, also outperforming the non-contextual dependency parsing baseline.

We also categorize Bambara, a Niger-Congo Bantu language spoken in Mali and surrounding countries, as Intermediate, relying mostly on the POS tagging results which follow similar patterns as Maltese and Narabizi\draftremove{(see  Table~\ref{tab:bambara})}. We note that the BambaraBERT that we trained achieves notably poor performance compared to the non-contextual baseline, a fact we attribute to the extremely low amount of available data (1000 sentences only). We also note that the non-contextual baseline is the best performing model for dependency parsing, which could also potentially classify Bambara as a ``Hard" language instead.

Our results in Wolof follow the same pattern. The non-contextual baseline achieves a 77.0 in LAS outperforming \mbert. 
However, \mlmtuning achieves the highest score of 77.9.



\paragraph{The importance of script}

We now turn our focus to Uralic languages. Finnish, Estonian, and Hungarian are high-resource representatives of this language family that are typically included in multilingual LMs, also having task-tuning data available in large quantities. However, for several smaller Uralic languages, task-tuning data are generally very scarce.

We report in {Table~\ref{tab:intermediate} the performance for two low-resource Uralic languages, namely Livvi and Erzya using 8-fold cross-validation, with each run only using around 700 training instances. \draftreplace{For both Livvi and Erzya, the multilingual model along with \mlmtuning achieves the best performance, outperforming the non-contextual baseline by more than 1.5 point for parsing and POS tagging. Additionally, we observe a}{Note the} striking difference between the parsing performance (LAS) of \mbert on Livvi, written with the Latin script, and on Erzya that uses the Cyrillic script. This suggests that the script could be playing a critical role when  transferring to those languages.} \draftadd{We explore this hypothesis in detail in section~\ref{sec:translit}.}

\subsection{Hard}
\label{sec:hard}

The last category of the hard unseen language is perhaps the most interesting one, as these languages are very hard to process. \draftreplace{All available large-scale language models are}{\mbert is} outperformed by non-contextual baselines as well as by monolingual language models trained from scratch on the available raw data. At the same time, \mlmtuning on the available raw data has a minimal impact on performance.

Uyghur, a Turkic language with about 10-15 million speakers in central Asia, is a prime example of a hard language for current models. In our experiments, outlined in Table~\ref{tab:hard_summary}, the non-contextual baseline outperforms all contextual variants, both monolingual and multilingual, in all the tasks with up to 20 points difference compared to \mbert for parsing. Additionally, the monolingual UyghurBERT trained on only 105K sentences outperforms \mbert even after \mlmtuning. 


We attribute this discrepancy to script differences: Uyghur uses the Perso-Arabic script, when the other Turkic languages that were part of \mbert pretraining use either the Latin (e.g. Turkish) or the Cyrillic script (e.g. Kazakh). 

Sorani Kurdish (also known as Central Kurdish) is a similarly hard language, mainly spoken in Iraqi Kurdistan by around 8 million speakers, which uses the Sorani alphabet, a variant of the Arabic script. We can solely evaluate on the NER task, \draftreplace{where the non-contextual baseline is the best model, achieving a 81.3 F1-score. The SoraniBert that we trained reaches 80.6 F1-score, while \mbert gets 70.4 F1-score.}{where the non-contextual baseline and the monolingual SoraniBERT perform similarly around 80.5 F1-score outperforming significantly \mbert which only reaches 70.4 in F1-score.} \mlmtuning on 380K sentences of Sorani texts improves \mbert performance to 75.6 F1-score, but it is still lagging behind the baseline. Our results in Sindhi follow the same pattern. The non-contextual baseline achieves a 51.4 F1-score outperforming with a large margin our language models (a monolingual SindhiBERT achieves an F1-score of 45.2, and \mbert is worse at 42.3).

\begin{table*}[t]
    \centering\footnotesize
    \begin{tabular}{c|ccc||cc}
    \toprule
         Model & \multicolumn{1}{c}{POS} & \multicolumn{1}{c}{LAS} & \multicolumn{1}{c}{NER} & Model & NER \\
    \midrule
    \multicolumn{4}{c||}{Uyghur (Arabic$\rightarrow$Latin)} & \multicolumn{2}{c}{Sorani (Arabic$\rightarrow$Latin)}\\
    UyghurBERT & 87.4$\rightarrow$86.2 & 57.3$\rightarrow$54.6 & 41.4$\rightarrow$41.7 
    & SoraniBERT & 80.6$\rightarrow$78.9 \\
    \mbert & 77.0$\rightarrow$87.9 & 45.7$\rightarrow$65.0 & 24.3$\rightarrow$35.7 &
    \mbert & 70.5$\rightarrow$77.8\\
    \mbertmlm & 77.3$\rightarrow$\textbf{89.8} & 48.9$\rightarrow$\textbf{66.8} & 34.7$\rightarrow$\textbf{55.2} & 
    \mbertmlm & 75.6$\rightarrow$\textbf{82.7}\\
    \midrule
    \multicolumn{4}{c||}{Buryat (Cyrillic$\rightarrow$Latin)} & \multicolumn{2}{c}{Meadow Mari (Cyrillic$\rightarrow$Latin)} \\
    BuryatBERT & 75.8$\rightarrow$75.8 & 31.4$\rightarrow$31.4 & -- & MariBERT & 44.0$\rightarrow$45.5\\
    \mbert & 83.9$\rightarrow$81.6 & 50.3$\rightarrow$45.8 & -- & \mbert & 55.2$\rightarrow$58.2\\
    \mbertmlm & \textbf{86.5}$\rightarrow$84.6 & \textbf{52.9}$\rightarrow$51.9 & -- & \mbertmlm & 57.6$\rightarrow$\textbf{65.9}\\
    \midrule
    
    \multicolumn{4}{c||}{Erzya (Cyrillic$\rightarrow$Latin)} & \multicolumn{2}{c}{Mingrelian (Georgian$\rightarrow$Latin)} \\
    ErzyaBERT & 84.4$\rightarrow$84.5 & 47.8$\rightarrow$47.8  & -- & MingrelianBERT & 42.0$\rightarrow$42.2\\
    \mbert & 89.3$\rightarrow$88.2 & 61.2$\rightarrow$58.3 & -- & \mbert & 53.6$\rightarrow$41.8\\
    \mbertmlm & \textbf{91.2}$\rightarrow$90.5 & \textbf{66.6}$\rightarrow$65.5 & -- & \mbertmlm & \textbf{68.4}$\rightarrow$62.6\\
    \bottomrule
    \end{tabular}
    \caption{Transliterating low-resource languages into the Latin script leads to significant improvements in languages like Uyghur, Sorani, and Meadow Mari. For languages like Erzya and Buryat transliteration, does not significantly influence results, while it does not help for Mingrelian. In all cases, \mbertmlm is the best approach.}
    \label{tab:translit}
\end{table*}

\section{Tackling Hard Languages with Multilingual Language Models}
\label{sec:tackling}

Our intermediate \draftadd{Uralic} language results provide initial supporting evidence for our argument on the importance of having pretrained LMs on languages with similar scripts, even for generally high-resource language families. \draftadd{Our hypothesis is that the script is a key element for \draftreplace{pretrained multilingual}{language} models to correctly process unseen languages.} 

To \draftreplace{verify our hypothesis on the importance of script}{test this hypothesis}, we assess the ability of \mbert to process an unseen language after transliterating it to another script \draftadd{present in the pretraining data}. 
We \draftreplace{focus our experiments}{experiment} on six languages belonging to four language families: Erzya, Bruyat and Meadow Mari (Uralic), Sorani Kurdish (Iranian, Indo-European), Uyghur (Turkic) and Mingrelian (Kartvelian). We apply the following transliterations: 
\begin{itemize}
    \item Erzya/Buryat/Mari: Cyrillic $\xrightarrow{}$ Latin Script 
    \item Uyghur: Arabic Script
    $\xrightarrow{}$ Latin Script
    \item Sorani: Arabic Script $\xrightarrow{}$ Latin Script
    \item Mingrelian: Georgian Script $\xrightarrow{}$ Latin Script
\end{itemize}

\subsection{Linguistically-motivated transliteration}

The strategy we used to transliterate the above-listed language is specific to the purpose of our experiments. Indeed, our goal is for the model to take advantage of the information it has learned during training on a related language written in the Latin script. The goal of our transliteration is therefore to transcribe each character in the source script, which we assume corresponds to a phoneme, into the most frequent (sometimes only) way this phoneme is rendered in the closest related language written in the Latin script, hereafter the target language. This process is not a transliteration strictly speaking, and it needs not be reversible. It is not a phonetization either, but rather a way to render the source language in a way that maximizes the similarity between the transliterated source language and the target language.

We have manually developed transliteration scripts for Uyghur and Sorani Kurdish, using respectively Turkish and Kurmanji Kurdish as target languages, only Turkish being one of the languages used to train mBERT. Note however that Turkish and Kurmanji Kurdish share a number of conventions for rendering phonemes in the Latin script (for instance, /\textesh/, rendered in English by ``sh'', is rendered in both languages by ``ş''; as a result, the Arabic letter ``\foreignlanguage{arabic}{ش}'', used in both languages, is rendered as ``ş'' by both our transliteration scripts). 
As for Erzya, Buryat and Mari, we used the readily available transliteration package {\em transliterate},\footnote{https://pypi.org/project/transliterate/} which performs a standard transliteration.\footnote{In future work, we intend to develop dedicated transliteration scripts using the strategy described above, and to compare the results obtained with it with those described here.} We used the Russian transliteration module, as it covers the Cyrillic script. Finally, for our control experiments on Mingrelian, we used the Georgian transliteration module from the same package.

\subsection{Transfer via Transliteration}
\label{sec:translit}
We train \mbert with \mlmtuning and \tasktuning as well as monolingual BERT model trained from scratch on the transliterated data. 
\draftadd{We also run controlled experiments  on high-resource languages written in the Latin script on which \mbert was pretrained on, namely Arabic, Japanese and Russian (reported in Table~\ref{tab:controlled_exp}).} 


Our results with and without transliteration are listed in Table~\ref{tab:translit}.
Transliteration for Sorani and Uyghur has a noticeable positive impact. For instance, transliterating Uyghur to Latin leads to an improvement of 16 points in parsing and 20 points in NER. For one of the low-resource Uralic languages, Meadow Mari, we observe an 8 F1-score points improvement on NER, while for other Uralic languages like Erzya the effect of transliteration is very minor. 
The only case where transliteration to the Latin script leads to a drop in performance for \mbert and \mbertmlm is Mingrelian.

We interpret our results as follows. When running \mlmtuning and \tasktuning, \mbert associates the target unseen language to a set of similar languages seen during pretraining based on the script. In consequence, \mbert is not able to associate a language to its related language if they are not written in the same script.  For instance, transliterating Uyghur enables \mbert to match it to Turkish, a language which accounts for a sizable portion of \mbert pretraining. In the case of Mingrelian, transliteration has the opposite effect: transliterating Mingrelian in the Latin script is harming the performance as \mbert is not able to associate it to Georgian which is seen during pretraining and uses the Georgian script.


This is further supported by our experiments  on high resource languages (cf. Table~\ref{tab:controlled_exp}). When transliterating pretrained languages such as Arabic, Russian or Japanese, \mbert is not able to compete with the performance reached when using the script seen during pretraining. Transliterating the Arabic script and the Cyrillic script to Latin does not automatically improve \mbert performance as it does for Sorani, Uyghur and Meadow Mari. For instance, transliterating Arabic to the Latin script leads to a drop in performance of 1.5, 4.1 and 6.9 points for POS tagging, parsing and NER respectively.\footnote{Details and complete results on these controlled experiments can be found in Appendix~\ref{sec:controlled}.}

Our findings are generally in line with previous work. Transliteration to English specifically~\cite{lin-etal-2016-leveraging,durrani-etal-2014-integrating} and named entity transliteration \cite{kundu-etal-2018-deep,grundkiewicz-heafield-2018-neural} has been proven useful for cross-lingual transfer in tasks like NER, entity linking~\cite{rijhwani2019zero}, morphological inflection~\cite{murikinati-etal-2020-transliteration}, and Machine Translation~\cite{amrhein-sennrich-2020-romanization}.

\begin{table}[t]
    \centering\footnotesize
    \scalebox{0.9}{ 
    \begin{tabular}{c|ccc}
    \toprule
        &\multicolumn{3}{c}{ Original Script $\rightarrow$ Latin Script}\\
         Model & \multicolumn{1}{c}{POS} & \multicolumn{1}{c}{LAS} & \multicolumn{1}{c}{NER} \\
         
    \midrule

    Arabic & 96.4 $\rightarrow$ 94.9 & 82.9 $\rightarrow$ 78.8  &  87.8 $\rightarrow$ 80.9  \\
    Russian & 98.1 $\rightarrow$ 96.0 & 88.4 $\rightarrow$ 84.5 &  88.1 $\rightarrow$ 86.0  \\
    Japanese & 97.4 $\rightarrow$ 95.7 & 88.5 $\rightarrow$ 86.9 & 61.5 $\rightarrow$ 55.6  \\
    
    \bottomrule
    
    \end{tabular}
    }
    \caption{\mbert \textsc{Task-Tuned} on high resource languages for POS tagging, parsing and NER.  We compare fine-tuning done on data written the original language script with fine-tuning done on Latin transliteration. In all cases, transliteration degrades downstream performance. 
    }
    \label{tab:controlled_exp}

\end{table}

The transliteration approach provides a viable path for rendering large pretrained models like \mbert useful for all languages of the world. Indeed, as reported in Table~\ref{tab:translit}, transliterating both Uyghur and Sorani leads to matching or outperforming the performance of non-contextual strong baselines and deliver usable models (e.g. +12.5 POS accuracy in Uyghur). 

\section{Discussion and Conclusion}
\label{sec:discussion}

Pretraining ever larger language models is a research direction that is currently receiving a lot of attention and resources from the NLP research community \cite{raffel2019exploring,brown2020language}. Still, a large majority of human languages are under-resourced making the development of monolingual language models very challenging in those settings. Another path is to build large scale multilingual language models.\footnote{Even though we explore a different research direction,  recent advances in small scale and domain specific language models  suggest such models could also have an important impact for those languages \cite{Micheli_et_al:2020:smallbertmodels}.}
However, such an approach faces the inherent zipfian structure of human languages, making the training of a single model to cover all languages an unfeasible solution \citep{conneau-etal-2020-unsupervised}. Reusing large scale pretrained language models for new unseen languages seems to be a more promising and reasonable solution from a cost-efficiency and environmental perspective \citep{strubell-etal-2019-energy}.

Recently, \citet{pfeiffer-etal-2020-mad} proposed to use adapter layers \citep{houlsby2019parameter} to build parameter efficient multilingual language models for unseen languages. However, this solution brings no significant improvement in the supervised setting, compared to a more simple Masked-Language Model finetuning. Furthermore, developing a language agnostic adaptation method is an unreasonable wish with regard to the large typological diversity of human languages. 

On the other hand, the promising vocabulary adaptation technique of~\citet{chau2020parsing} which leads to good dependency parsing results on unseen languages when combined with task-tuning has so far been tested only on Latin script languages (Singlish and Maltese). We expect that it will be orthogonal to our transliteration approach, but we leave for future work the study of its applicability and efficacy on more languages and tasks.

In this context, we bring empirical evidence to assess the efficiency of language models pretraining and adaptation methods on 15 low-resource and typologically diverse unseen languages.  Our results show that the ``Hard" languages are currently out-of-the-scope of any currently available language models and are therefore left outside of the current NLP progress. By focusing on those, we find that this challenge is mostly due to the script. Transliterating them to a script that is used by a related higher resource language on which the language model has been pretrained on leads to large improvements in downstream performance. 
Our results shed some new light on the importance of the script in multilingual pretrained models. While previous work suggests that multilingual language models could transfer efficiently across scripts in zero-shot settings \citep{pires2019multilingual,karthikeyan2019cross}, our results show that such cross-script transfer is possible only if the model has seen related languages in the same script during pretraining. 

Our work paves the way for a better understanding of the mechanics at play in cross-language transfer learning in low-resource scenarios. We strongly believe that our method can contribute to bootstrapping NLP resources and tools for low-resource languages, thereby favoring the emergence of NLP ecosystems for languages currently under-served by the NLP community.

\section*{Acknowledgments}

The Inria authors were partly funded by two French Research National agency  projects, namely projects PARSITI (ANR-16-CE33-0021) and SoSweet (ANR-15-CE38-0011), as well as by Benoit Sagot's chair in the PRAIRIE institute as part of the ``Investissements d’avenir'' programme under the reference \mbox{ANR-19-P3IA-0001}. 
Antonios Anastasopoulos is generously supported by NSF Award 2040926 and is also thankful to Graham Neubig for very insightful initial discussions on this research direction.


\newpage
\textcolor{white}{.}
\newpage

\bibliography{tacl2018}

\begin{thebibliography}{52}
\expandafter\ifx\csname natexlab\endcsname\relax\def\natexlab#1{#1}\fi

\bibitem[{\'{A}cs(2019)}]{acs2019exploring}
Judit \'{A}cs. 2019.
\newblock Exploring bert's vocabulary.
\newblock Http://juditacs.github.io/2019/02/19/bert-tokenization-stats.html.

\bibitem[{Ammar et~al.(2016)Ammar, Mulcaire, Tsvetkov, Lample, Dyer, and
  Smith}]{ammar2016massively}
Waleed Ammar, George Mulcaire, Yulia Tsvetkov, Guillaume Lample, Chris Dyer,
  and Noah~A Smith. 2016.
\newblock Massively multilingual word embeddings.
\newblock {arXiv}:1602.01925.

\bibitem[{Antoun et~al.(2020)Antoun, Baly, and Hajj}]{antoun2020arabert}
Wissam Antoun, Fady Baly, and Hazem Hajj. 2020.
\newblock Arabert: Transformer-based model for arabic language understanding.
\newblock {arXiv}:2003.00104.

\bibitem[{Biemann et~al.(2007)Biemann, Heyer, Quasthoff, and
  Richter}]{biemann2007leipzig}
Chris Biemann, Gerhard Heyer, Uwe Quasthoff, and Matthias Richter. 2007.
\newblock The leipzig corpora collection-monolingual corpora of standard size.
\newblock \emph{Proceedings of Corpus Linguistic}, 2007.

\bibitem[{Brown et~al.(2020)Brown, Mann, Ryder, Subbiah, Kaplan, Dhariwal,
  Neelakantan, Shyam, Sastry, Askell et~al.}]{brown2020language}
Tom~B Brown, Benjamin Mann, Nick Ryder, Melanie Subbiah, Jared Kaplan, Prafulla
  Dhariwal, Arvind Neelakantan, Pranav Shyam, Girish Sastry, Amanda Askell,
  et~al. 2020.
\newblock Language models are few-shot learners.
\newblock {arXiv}:2005.14165.

\bibitem[{Caron et~al.(2019)Caron, Courtin, Gerdes, and
  Kahane}]{caron2019surface}
Bernard Caron, Marine Courtin, Kim Gerdes, and Sylvain Kahane. 2019.
\newblock A surface-syntactic ud treebank for naija.
\newblock In \emph{Proceedings of the 18th International Workshop on Treebanks
  and Linguistic Theories (TLT, SyntaxFest 2019)}, pages 13--24.

\bibitem[{Cañete et~al.(2020)Cañete, Chaperon, Fuentes, and
  Pérez}]{canete-2020-beto}
José Cañete, Gabriel Chaperon, Rodrigo Fuentes, and Jorge Pérez. 2020.
\newblock \href {https://users.dcc.uchile.cl/~jperez/papers/pml4dc2020.pdf}
  {Spanish pre-trained bert model and evaluation data}.
\newblock In \emph{PML4DC at ICLR 2020}.

\bibitem[{Chau et~al.(2020)Chau, Lin, and Smith}]{chau2020parsing}
Ethan~C Chau, Lucy~H Lin, and Noah~A Smith. 2020.
\newblock Parsing with multilingual bert, a small corpus, and a small treebank.
\newblock {arXiv}:2009.14124.

\bibitem[{Conneau et~al.(2020)Conneau, Khandelwal, Goyal, Chaudhary, Wenzek,
  Guzm{\'a}n, Grave, Ott, Zettlemoyer, and
  Stoyanov}]{conneau-etal-2020-unsupervised}
Alexis Conneau, Kartikay Khandelwal, Naman Goyal, Vishrav Chaudhary, Guillaume
  Wenzek, Francisco Guzm{\'a}n, Edouard Grave, Myle Ott, Luke Zettlemoyer, and
  Veselin Stoyanov. 2020.
\newblock \href {https://www.aclweb.org/anthology/2020.acl-main.747}
  {Unsupervised cross-lingual representation learning at scale}.
\newblock In \emph{Proceedings of the 58th Annual Meeting of the Association
  for Computational Linguistics}, pages 8440--8451, Online. Association for
  Computational Linguistics.

\bibitem[{Devlin et~al.(2019)Devlin, Chang, Lee, and
  Toutanova}]{devlin-etal-2019-bert}
Jacob Devlin, Ming-Wei Chang, Kenton Lee, and Kristina Toutanova. 2019.
\newblock \href {https://doi.org/10.18653/v1/N19-1423} {{BERT}: Pre-training of
  deep bidirectional transformers for language understanding}.
\newblock In \emph{Proceedings of the 2019 Conference of the North {A}merican
  Chapter of the Association for Computational Linguistics: Human Language
  Technologies, Volume 1 (Long and Short Papers)}, pages 4171--4186,
  Minneapolis, Minnesota. Association for Computational Linguistics.

\bibitem[{Dozat and Manning(2016)}]{dozat2016deep}
Timothy Dozat and Christopher~D Manning. 2016.
\newblock Deep biaffine attention for neural dependency parsing.
\newblock {arXiv}:1611.01734.

\bibitem[{Durrani et~al.(2014)Durrani, Sajjad, Hoang, and
  Koehn}]{durrani-etal-2014-integrating}
Nadir Durrani, Hassan Sajjad, Hieu Hoang, and Philipp Koehn. 2014.
\newblock \href {https://doi.org/10.3115/v1/E14-4029} {Integrating an
  unsupervised transliteration model into statistical machine translation}.
\newblock In \emph{Proceedings of the 14th Conference of the {E}uropean Chapter
  of the Association for Computational Linguistics, volume 2: Short Papers},
  pages 148--153, Gothenburg, Sweden. Association for Computational
  Linguistics.

\bibitem[{Grundkiewicz and Heafield(2018)}]{grundkiewicz-heafield-2018-neural}
Roman Grundkiewicz and Kenneth Heafield. 2018.
\newblock \href {https://doi.org/10.18653/v1/W18-2413} {Neural machine
  translation techniques for named entity transliteration}.
\newblock In \emph{Proceedings of the Seventh Named Entities Workshop}, pages
  89--94, Melbourne, Australia. Association for Computational Linguistics.

\bibitem[{Han and Eisenstein(2019)}]{han-eisenstein-2019-unsupervised}
Xiaochuang Han and Jacob Eisenstein. 2019.
\newblock \href {https://doi.org/10.18653/v1/D19-1433} {Unsupervised domain
  adaptation of contextualized embeddings for sequence labeling}.
\newblock In \emph{Proceedings of the 2019 Conference on Empirical Methods in
  Natural Language Processing and the 9th International Joint Conference on
  Natural Language Processing (EMNLP-IJCNLP)}, pages 4238--4248, Hong Kong,
  China. Association for Computational Linguistics.

\bibitem[{Hochreiter and Schmidhuber(1997)}]{hochreiter1997long}
Sepp Hochreiter and J{\"u}rgen Schmidhuber. 1997.
\newblock Long short-term memory.
\newblock \emph{Neural computation}, 9(8):1735--1780.

\bibitem[{Houlsby et~al.(2019)Houlsby, Giurgiu, Jastrzebski, Morrone,
  De~Laroussilhe, Gesmundo, Attariyan, and Gelly}]{houlsby2019parameter}
Neil Houlsby, Andrei Giurgiu, Stanislaw Jastrzebski, Bruna Morrone, Quentin
  De~Laroussilhe, Andrea Gesmundo, Mona Attariyan, and Sylvain Gelly. 2019.
\newblock Parameter-efficient transfer learning for {NLP}.
\newblock {arXiv}:1902.00751.

\bibitem[{Joshi et~al.(2020)Joshi, Santy, Budhiraja, Bali, and
  Choudhury}]{joshi2020state}
Pratik Joshi, Sebastin Santy, Amar Budhiraja, Kalika Bali, and Monojit
  Choudhury. 2020.
\newblock \href {https://doi.org/10.18653/v1/2020.acl-main.560} {The state and
  fate of linguistic diversity and inclusion in the {NLP} world}.
\newblock In \emph{Proceedings of the 58th Annual Meeting of the Association
  for Computational Linguistics}, pages 6282--6293, Online. Association for
  Computational Linguistics.

\bibitem[{K et~al.(2020)K, Wang, Mayhew, and Roth}]{K2020Cross-Lingual}
Karthikeyan K, Zihan Wang, Stephen Mayhew, and Dan Roth. 2020.
\newblock \href {https://openreview.net/forum?id=HJeT3yrtDr} {Cross-lingual
  ability of multilingual bert: An empirical study}.
\newblock In \emph{International Conference on Learning Representations}.

\bibitem[{Kingma and Ba(2014)}]{kingma2014adam}
Diederik~P Kingma and Jimmy Ba. 2014.
\newblock Adam: A method for stochastic optimization.
\newblock {arXiv}:1412.6980.

\bibitem[{Kondratyuk and Straka(2019)}]{kondratyuk-straka-2019-75}
Dan Kondratyuk and Milan Straka. 2019.
\newblock \href {https://doi.org/10.18653/v1/D19-1279} {75 languages, 1 model:
  Parsing universal dependencies universally}.
\newblock In \emph{Proceedings of the 2019 Conference on Empirical Methods in
  Natural Language Processing and the 9th International Joint Conference on
  Natural Language Processing (EMNLP-IJCNLP)}, pages 2779--2795, Hong Kong,
  China. Association for Computational Linguistics.

\bibitem[{Kudo and Richardson(2018)}]{kudo-richardson-2018-sentencepiece}
Taku Kudo and John Richardson. 2018.
\newblock \href {https://doi.org/10.18653/v1/D18-2012} {{S}entence{P}iece: A
  simple and language independent subword tokenizer and detokenizer for neural
  text processing}.
\newblock In \emph{Proceedings of the 2018 Conference on Empirical Methods in
  Natural Language Processing: System Demonstrations}, pages 66--71, Brussels,
  Belgium. Association for Computational Linguistics.

\bibitem[{Kundu et~al.(2018)Kundu, Paul, and Pal}]{kundu-etal-2018-deep}
Soumyadeep Kundu, Sayantan Paul, and Santanu Pal. 2018.
\newblock \href {https://doi.org/10.18653/v1/W18-2411} {A deep learning based
  approach to transliteration}.
\newblock In \emph{Proceedings of the Seventh Named Entities Workshop}, pages
  79--83, Melbourne, Australia. Association for Computational Linguistics.

\bibitem[{Kuratov and Arkhipov(2019)}]{kuratov-2019-rubert}
Yuri Kuratov and Mikhail Arkhipov. 2019.
\newblock \href {http://arxiv.org/abs/1905.07213} {Adaptation of deep
  bidirectional multilingual transformers for russian language}.
\newblock \emph{CoRR}, abs/1905.07213.

\bibitem[{Lin et~al.(2016)Lin, Pan, Deri, Ji, and
  Knight}]{lin-etal-2016-leveraging}
Ying Lin, Xiaoman Pan, Aliya Deri, Heng Ji, and Kevin Knight. 2016.
\newblock \href {https://www.aclweb.org/anthology/W16-2701} {Leveraging entity
  linking and related language projection to improve name transliteration}.
\newblock In \emph{Proceedings of the Sixth Named Entity Workshop}, pages
  1--10, Berlin, Germany. Association for Computational Linguistics.

\bibitem[{Liu et~al.(2019)Liu, Ott, Goyal, Du, Joshi, Chen, Levy, Lewis,
  Zettlemoyer, and Stoyanov}]{liu2019roberta}
Yinhan Liu, Myle Ott, Naman Goyal, Jingfei Du, Mandar Joshi, Danqi Chen, Omer
  Levy, Mike Lewis, Luke Zettlemoyer, and Veselin Stoyanov. 2019.
\newblock {RoBERTa}: A robustly optimized {BERT} pretraining approach.
\newblock {arXiv}:1907.11692.

\bibitem[{Martin et~al.(2020{\natexlab{a}})Martin, Muller, Ortiz~Su{\'a}rez,
  Dupont, Romary, de~la Clergerie, Seddah, and
  Sagot}]{martin-etal-2020-camembert}
Louis Martin, Benjamin Muller, Pedro~Javier Ortiz~Su{\'a}rez, Yoann Dupont,
  Laurent Romary, {\'E}ric de~la Clergerie, Djam{\'e} Seddah, and Beno{\^\i}t
  Sagot. 2020{\natexlab{a}}.
\newblock \href {https://www.aclweb.org/anthology/2020.acl-main.645}
  {{C}amem{BERT}: a tasty {F}rench language model}.
\newblock In \emph{Proceedings of the 58th Annual Meeting of the Association
  for Computational Linguistics}, pages 7203--7219, Online. Association for
  Computational Linguistics.

\bibitem[{Martin et~al.(2020{\natexlab{b}})Martin, Muller, Ortiz~Su{\'a}rez,
  Dupont, Romary, de~la Clergerie, Seddah, and Sagot}]{martin-2019-camembert}
Louis Martin, Benjamin Muller, Pedro~Javier Ortiz~Su{\'a}rez, Yoann Dupont,
  Laurent Romary, {\'E}ric~Villemonte de~la Clergerie, Djam{\'e} Seddah, and
  Beno{\^\i}t Sagot. 2020{\natexlab{b}}.
\newblock \href {https://arxiv.org/abs/1911.03894} {Camembert: a tasty french
  language model}.
\newblock In \emph{Proceedings of the 58th Annual Meeting of the Association
  for Computational Linguistics}.

\bibitem[{Micheli et~al.(2020)Micheli, d'Hoffschmidt, and
  Fleuret}]{Micheli_et_al:2020:smallbertmodels}
Vincent Micheli, Martin d'Hoffschmidt, and François Fleuret. 2020.
\newblock \href {http://arxiv.org/abs/arXiv:2010.03813} {On the importance of
  pre-training data volume for compact language models}.

\bibitem[{Mikolov et~al.(2013)Mikolov, Sutskever, Chen, Corrado, and
  Dean}]{mikolov2013distributed}
Tomas Mikolov, Ilya Sutskever, Kai Chen, Greg~S Corrado, and Jeff Dean. 2013.
\newblock Distributed representations of words and phrases and their
  compositionality.
\newblock In \emph{Advances in neural information processing systems}, pages
  3111--3119.

\bibitem[{Murikinati et~al.(2020)Murikinati, Anastasopoulos, and
  Neubig}]{murikinati-etal-2020-transliteration}
Nikitha Murikinati, Antonios Anastasopoulos, and Graham Neubig. 2020.
\newblock \href {https://doi.org/10.18653/v1/2020.sigmorphon-1.22}
  {Transliteration for cross-lingual morphological inflection}.
\newblock In \emph{Proceedings of the 17th SIGMORPHON Workshop on Computational
  Research in Phonetics, Phonology, and Morphology}, pages 189--197, Online.
  Association for Computational Linguistics.

\bibitem[{Nivre et~al.(2016)Nivre, De~Marneffe, Ginter, Goldberg, Hajic,
  Manning, McDonald, Petrov, Pyysalo, Silveira et~al.}]{nivre2016universal}
Joakim Nivre, Marie-Catherine De~Marneffe, Filip Ginter, Yoav Goldberg, Jan
  Hajic, Christopher~D Manning, Ryan McDonald, Slav Petrov, Sampo Pyysalo,
  Natalia Silveira, et~al. 2016.
\newblock Universal dependencies v1: A multilingual treebank collection.
\newblock In \emph{Proceedings of the Tenth International Conference on
  Language Resources and Evaluation (LREC'16)}, pages 1659--1666.

\bibitem[{{Ortiz Su{\'a}rez} et~al.(2019){Ortiz Su{\'a}rez}, Sagot, and
  Romary}]{OrtizSuarezSagotRomary2019}
Pedro~Javier {Ortiz Su{\'a}rez}, Beno{\^i}t Sagot, and Laurent Romary. 2019.
\newblock \href {https://doi.org/10.14618/ids-pub-9021} {Asynchronous pipelines
  for processing huge corpora on medium to low resource infrastructures}.
\newblock In \emph{Proceedings of the Workshop on Challenges in the Management
  of Large Corpora (CMLC-7) 2019. Cardiff, 22nd July 2019}, pages 9 -- 16,
  Mannheim. Leibniz-Institut f{\"u}r Deutsche Sprache.

\bibitem[{Pan et~al.(2017)Pan, Zhang, May, Nothman, Knight, and
  Ji}]{pan-etal-2017-cross}
Xiaoman Pan, Boliang Zhang, Jonathan May, Joel Nothman, Kevin Knight, and Heng
  Ji. 2017.
\newblock \href {https://doi.org/10.18653/v1/P17-1178} {Cross-lingual name
  tagging and linking for 282 languages}.
\newblock In \emph{Proceedings of the 55th Annual Meeting of the Association
  for Computational Linguistics (Volume 1: Long Papers)}, pages 1946--1958,
  Vancouver, Canada. Association for Computational Linguistics.

\bibitem[{Peters et~al.(2018)Peters, Neumann, Iyyer, Gardner, Clark, Lee, and
  Zettlemoyer}]{Peters:2018}
Matthew~E. Peters, Mark Neumann, Mohit Iyyer, Matt Gardner, Christopher Clark,
  Kenton Lee, and Luke Zettlemoyer. 2018.
\newblock Deep contextualized word representations.
\newblock In \emph{Proc. of NAACL}.

\bibitem[{Pfeiffer et~al.(2020)Pfeiffer, Vuli{\'c}, Gurevych, and
  Ruder}]{pfeiffer2020mad}
Jonas Pfeiffer, Ivan Vuli{\'c}, Iryna Gurevych, and Sebastian Ruder. 2020.
\newblock Mad-x: An adapter-based framework for multi-task cross-lingual
  transfer.
\newblock {arXiv}:2005.00052.

\bibitem[{Pires et~al.(2019)Pires, Schlinger, and
  Garrette}]{pires2019multilingual}
Telmo Pires, Eva Schlinger, and Dan Garrette. 2019.
\newblock How multilingual is multilingual bert?
\newblock In \emph{Proceedings of the 57th Annual Meeting of the Association
  for Computational Linguistics}, pages 4996--5001.

\bibitem[{Qi et~al.(2020)Qi, Zhang, Zhang, Bolton, and Manning}]{qi2020stanza}
Peng Qi, Yuhao Zhang, Yuhui Zhang, Jason Bolton, and Christopher~D Manning.
  2020.
\newblock Stanza: A python natural language processing toolkit for many human
  languages.
\newblock {arXiv}:2003.07082.

\bibitem[{Raffel et~al.(2019)Raffel, Shazeer, Roberts, Lee, Narang, Matena,
  Zhou, Li, and Liu}]{raffel2019exploring}
Colin Raffel, Noam Shazeer, Adam Roberts, Katherine Lee, Sharan Narang, Michael
  Matena, Yanqi Zhou, Wei Li, and Peter~J Liu. 2019.
\newblock Exploring the limits of transfer learning with a unified text-to-text
  transformer.
\newblock {arXiv}:1910.10683.

\bibitem[{Rahimi et~al.(2019)Rahimi, Li, and Cohn}]{rahimi-etal-2019-massively}
Afshin Rahimi, Yuan Li, and Trevor Cohn. 2019.
\newblock \href {https://doi.org/10.18653/v1/P19-1015} {Massively multilingual
  transfer for {NER}}.
\newblock In \emph{Proceedings of the 57th Annual Meeting of the Association
  for Computational Linguistics}, pages 151--164, Florence, Italy. Association
  for Computational Linguistics.

\bibitem[{Rijhwani et~al.(2019)Rijhwani, Xie, Neubig, and
  Carbonell}]{rijhwani2019zero}
Shruti Rijhwani, Jiateng Xie, Graham Neubig, and Jaime Carbonell. 2019.
\newblock Zero-shot neural transfer for cross-lingual entity linking.
\newblock In \emph{Proceedings of the AAAI Conference on Artificial
  Intelligence}, volume~33, pages 6924--6931.

\bibitem[{Schmidt and Wiegand(2017)}]{schmidt-wiegand-2017-survey}
Anna Schmidt and Michael Wiegand. 2017.
\newblock \href {https://doi.org/10.18653/v1/W17-1101} {A survey on hate speech
  detection using natural language processing}.
\newblock In \emph{Proceedings of the Fifth International Workshop on Natural
  Language Processing for Social Media}, pages 1--10, Valencia, Spain.
  Association for Computational Linguistics.

\bibitem[{Schweter(2020)}]{schweter-2020-berturk}
Stefan Schweter. 2020.
\newblock \href {https://doi.org/10.5281/zenodo.3770924} {Berturk - bert models
  for turkish}.

\bibitem[{Seddah et~al.(2020)Seddah, Essaidi, Fethi, Futeral, Muller,
  Ortiz~Su{\'a}rez, Sagot, and Srivastava}]{seddah-etal-2020-building}
Djam{\'e} Seddah, Farah Essaidi, Amal Fethi, Matthieu Futeral, Benjamin Muller,
  Pedro~Javier Ortiz~Su{\'a}rez, Beno{\^\i}t Sagot, and Abhishek Srivastava.
  2020.
\newblock \href {https://doi.org/10.18653/v1/2020.acl-main.107} {Building a
  user-generated content {N}orth-{A}frican {A}rabizi treebank: Tackling hell}.
\newblock In \emph{Proceedings of the 58th Annual Meeting of the Association
  for Computational Linguistics}, pages 1139--1150, Online. Association for
  Computational Linguistics.

\bibitem[{Stecklow(2018)}]{myanmar2018}
Steve Stecklow. 2018.
\newblock {Why Facebook is losing the war on hate speech in Myanmar, Reuters}.
\newblock
  \url{https://www.reuters.com/investigates/special-report/myanmar-facebook-hate/
  }.

\bibitem[{Straka(2018)}]{straka2018udpipe}
Milan Straka. 2018.
\newblock Udpipe 2.0 prototype at conll 2018 ud shared task.
\newblock In \emph{Proceedings of the CoNLL 2018 Shared Task: Multilingual
  Parsing from Raw Text to Universal Dependencies}, pages 197--207.

\bibitem[{Strubell et~al.(2019)Strubell, Ganesh, and
  McCallum}]{strubell-etal-2019-energy}
Emma Strubell, Ananya Ganesh, and Andrew McCallum. 2019.
\newblock \href {https://doi.org/10.18653/v1/P19-1355} {Energy and policy
  considerations for deep learning in {NLP}}.
\newblock In \emph{Proceedings of the 57th Annual Meeting of the Association
  for Computational Linguistics}, pages 3645--3650, Florence, Italy.
  Association for Computational Linguistics.

\bibitem[{Su{\'a}rez et~al.(2020)Su{\'a}rez, Romary, and
  Sagot}]{suarez2020monolingual}
Pedro~Ortiz Su{\'a}rez, Laurent Romary, and Beno{\^\i}t Sagot. 2020.
\newblock A monolingual approach to contextualized word embeddings for
  mid-resource languages.
\newblock {arXiv}:2006.06202.

\bibitem[{de~Vries et~al.(2019)de~Vries, van Cranenburgh, Bisazza, Caselli, van
  Noord, and Nissim}]{vries-2019-bertje}
Wietse de~Vries, Andreas van Cranenburgh, Arianna Bisazza, Tommaso Caselli,
  Gertjan van Noord, and Malvina Nissim. 2019.
\newblock \href {http://arxiv.org/abs/1912.09582} {Bertje: {A} dutch {BERT}
  model}.
\newblock \emph{CoRR}, abs/1912.09582.

\bibitem[{Wang et~al.(2019)Wang, Mayhew, Roth et~al.}]{wang2019cross}
Zihan Wang, Stephen Mayhew, Dan Roth, et~al. 2019.
\newblock Cross-lingual ability of multilingual bert: An empirical study.
\newblock {arXiv}:1912.07840.

\bibitem[{Wolf et~al.(2019)Wolf, Debut, Sanh, Chaumond, Delangue, Moi, Cistac,
  Rault, Louf, Funtowicz et~al.}]{wolf2019transformers}
Thomas Wolf, Lysandre Debut, Victor Sanh, Julien Chaumond, Clement Delangue,
  Anthony Moi, Pierric Cistac, Tim Rault, R{\'e}mi Louf, Morgan Funtowicz,
  et~al. 2019.
\newblock Transformers: State-of-the-art natural language processing.
\newblock \emph{arXiv preprint arXiv:1910.03771}.

\bibitem[{Wu and Dredze(2020)}]{wu2020all}
Shijie Wu and Mark Dredze. 2020.
\newblock Are all languages created equal in multilingual bert?
\newblock \emph{arXiv preprint arXiv:2005.09093}.

\bibitem[{Zeman and Haji{\v{c}}(2018)}]{conll-2018-conll}
Daniel Zeman and Jan Haji{\v{c}}, editors. 2018.
\newblock \href {https://www.aclweb.org/anthology/K18-2000} {\emph{Proceedings
  of the {C}o{NLL} 2018 Shared Task: Multilingual Parsing from Raw Text to
  Universal Dependencies}}. Association for Computational Linguistics,
  Brussels, Belgium.

\end{thebibliography}


\begin{thebibliography}{53}
\expandafter\ifx\csname natexlab\endcsname\relax\def\natexlab#1{#1}\fi

\bibitem[{\'{A}cs(2019)}]{acs2019exploring}
Judit \'{A}cs. 2019.
\newblock Exploring {BERT}'s vocabulary.
\newblock Http://juditacs.github.io/2019/02/19/bert-tokenization-stats.html.

\bibitem[{Ammar et~al.(2016)Ammar, Mulcaire, Tsvetkov, Lample, Dyer, and
  Smith}]{ammar2016massively}
Waleed Ammar, George Mulcaire, Yulia Tsvetkov, Guillaume Lample, Chris Dyer,
  and Noah~A Smith. 2016.
\newblock Massively multilingual word embeddings.
\newblock {arXiv}:1602.01925.

\bibitem[{Amrhein and Sennrich(2020)}]{amrhein-sennrich-2020-romanization}
Chantal Amrhein and Rico Sennrich. 2020.
\newblock \href {https://www.aclweb.org/anthology/2020.findings-emnlp.223} {On
  {R}omanization for model transfer between scripts in neural machine
  translation}.
\newblock In \emph{Findings of the Association for Computational Linguistics:
  EMNLP 2020}, pages 2461--2469, Online. Association for Computational
  Linguistics.

\bibitem[{Antoun et~al.(2020)Antoun, Baly, and Hajj}]{antoun2020arabert}
Wissam Antoun, Fady Baly, and Hazem Hajj. 2020.
\newblock \href {https://www.aclweb.org/anthology/2020.osact-1.2} {{A}ra{BERT}:
  Transformer-based model for {A}rabic language understanding}.
\newblock In \emph{Proceedings of the 4th Workshop on Open-Source Arabic
  Corpora and Processing Tools, with a Shared Task on Offensive Language
  Detection}, pages 9--15, Marseille, France. European Language Resource
  Association.

\bibitem[{Biemann et~al.(2007)Biemann, Heyer, Quasthoff, and
  Richter}]{biemann2007leipzig}
Chris Biemann, Gerhard Heyer, Uwe Quasthoff, and Matthias Richter. 2007.
\newblock The {L}eipzig {C}orpora collection-monolingual corpora of standard
  size.
\newblock \emph{Proceedings of Corpus Linguistic}, 2007.

\bibitem[{Brown et~al.(2020)Brown, Mann, Ryder, Subbiah, Kaplan, Dhariwal,
  Neelakantan, Shyam, Sastry, Askell et~al.}]{brown2020language}
Tom~B Brown, Benjamin Mann, Nick Ryder, Melanie Subbiah, Jared Kaplan, Prafulla
  Dhariwal, Arvind Neelakantan, Pranav Shyam, Girish Sastry, Amanda Askell,
  et~al. 2020.
\newblock Language models are few-shot learners.
\newblock {arXiv}:2005.14165.

\bibitem[{Caron et~al.(2019)Caron, Courtin, Gerdes, and
  Kahane}]{caron2019surface}
Bernard Caron, Marine Courtin, Kim Gerdes, and Sylvain Kahane. 2019.
\newblock A surface-syntactic {UD} treebank for {N}aija.
\newblock In \emph{Proceedings of the 18th International Workshop on Treebanks
  and Linguistic Theories (TLT, SyntaxFest 2019)}, pages 13--24.

\bibitem[{Cañete et~al.(2020)Cañete, Chaperon, Fuentes, and
  Pérez}]{canete-2020-beto}
José Cañete, Gabriel Chaperon, Rodrigo Fuentes, and Jorge Pérez. 2020.
\newblock \href {https://users.dcc.uchile.cl/~jperez/papers/pml4dc2020.pdf}
  {Spanish pre-trained {BERT} model and evaluation data}.
\newblock In \emph{PML4DC at ICLR 2020}.

\bibitem[{Chau et~al.(2020)Chau, Lin, and Smith}]{chau2020parsing}
Ethan~C. Chau, Lucy~H. Lin, and Noah~A. Smith. 2020.
\newblock \href {https://doi.org/10.18653/v1/2020.findings-emnlp.118} {Parsing
  with multilingual {BERT}, a small corpus, and a small treebank}.
\newblock In \emph{Findings of the Association for Computational Linguistics:
  EMNLP 2020}, pages 1324--1334, Online. Association for Computational
  Linguistics.

\bibitem[{Conneau et~al.(2020)Conneau, Khandelwal, Goyal, Chaudhary, Wenzek,
  Guzm{\'a}n, Grave, Ott, Zettlemoyer, and
  Stoyanov}]{conneau-etal-2020-unsupervised}
Alexis Conneau, Kartikay Khandelwal, Naman Goyal, Vishrav Chaudhary, Guillaume
  Wenzek, Francisco Guzm{\'a}n, Edouard Grave, Myle Ott, Luke Zettlemoyer, and
  Veselin Stoyanov. 2020.
\newblock \href {https://www.aclweb.org/anthology/2020.acl-main.747}
  {Unsupervised cross-lingual representation learning at scale}.
\newblock In \emph{Proceedings of the 58th Annual Meeting of the Association
  for Computational Linguistics}, pages 8440--8451, Online. Association for
  Computational Linguistics.

\bibitem[{Devlin et~al.(2019)Devlin, Chang, Lee, and
  Toutanova}]{devlin-etal-2019-bert}
Jacob Devlin, Ming-Wei Chang, Kenton Lee, and Kristina Toutanova. 2019.
\newblock \href {https://doi.org/10.18653/v1/N19-1423} {{BERT}: Pre-training of
  deep bidirectional transformers for language understanding}.
\newblock In \emph{Proceedings of the 2019 Conference of the North {A}merican
  Chapter of the Association for Computational Linguistics: Human Language
  Technologies, Volume 1 (Long and Short Papers)}, pages 4171--4186,
  Minneapolis, Minnesota. Association for Computational Linguistics.

\bibitem[{Dozat and Manning(2017)}]{dozat2016deep}
Timothy Dozat and Christopher~D. Manning. 2017.
\newblock \href {https://openreview.net/forum?id=Hk95PK9le} {Deep biaffine
  attention for neural dependency parsing}.
\newblock In \emph{5th International Conference on Learning Representations,
  {ICLR} 2017, Toulon, France, April 24-26, 2017, Conference Track
  Proceedings}. OpenReview.net.

\bibitem[{Durrani et~al.(2014)Durrani, Sajjad, Hoang, and
  Koehn}]{durrani-etal-2014-integrating}
Nadir Durrani, Hassan Sajjad, Hieu Hoang, and Philipp Koehn. 2014.
\newblock \href {https://doi.org/10.3115/v1/E14-4029} {Integrating an
  unsupervised transliteration model into statistical machine translation}.
\newblock In \emph{Proceedings of the 14th Conference of the {E}uropean Chapter
  of the Association for Computational Linguistics, volume 2: Short Papers},
  pages 148--153, Gothenburg, Sweden. Association for Computational
  Linguistics.

\bibitem[{Grundkiewicz and Heafield(2018)}]{grundkiewicz-heafield-2018-neural}
Roman Grundkiewicz and Kenneth Heafield. 2018.
\newblock \href {https://doi.org/10.18653/v1/W18-2413} {Neural machine
  translation techniques for named entity transliteration}.
\newblock In \emph{Proceedings of the Seventh Named Entities Workshop}, pages
  89--94, Melbourne, Australia. Association for Computational Linguistics.

\bibitem[{Hammarstr{\"o}m(2016)}]{hammarstrom2016linguistic}
Harald Hammarstr{\"o}m. 2016.
\newblock Linguistic diversity and language evolution.
\newblock \emph{Journal of Language Evolution}, 1(1):19--29.

\bibitem[{Han and Eisenstein(2019)}]{han-eisenstein-2019-unsupervised}
Xiaochuang Han and Jacob Eisenstein. 2019.
\newblock \href {https://doi.org/10.18653/v1/D19-1433} {Unsupervised domain
  adaptation of contextualized embeddings for sequence labeling}.
\newblock In \emph{Proceedings of the 2019 Conference on Empirical Methods in
  Natural Language Processing and the 9th International Joint Conference on
  Natural Language Processing (EMNLP-IJCNLP)}, pages 4238--4248, Hong Kong,
  China. Association for Computational Linguistics.

\bibitem[{Hochreiter and Schmidhuber(1997)}]{hochreiter1997long}
Sepp Hochreiter and J{\"u}rgen Schmidhuber. 1997.
\newblock Long short-term memory.
\newblock \emph{Neural computation}, 9(8):1735--1780.

\bibitem[{Houlsby et~al.(2019)Houlsby, Giurgiu, Jastrzebski, Morrone,
  De~Laroussilhe, Gesmundo, Attariyan, and Gelly}]{houlsby2019parameter}
Neil Houlsby, Andrei Giurgiu, Stanislaw Jastrzebski, Bruna Morrone, Quentin
  De~Laroussilhe, Andrea Gesmundo, Mona Attariyan, and Sylvain Gelly. 2019.
\newblock Parameter-efficient transfer learning for {NLP}.
\newblock {arXiv}:1902.00751.

\bibitem[{Joshi et~al.(2020)Joshi, Santy, Budhiraja, Bali, and
  Choudhury}]{joshi2020state}
Pratik Joshi, Sebastin Santy, Amar Budhiraja, Kalika Bali, and Monojit
  Choudhury. 2020.
\newblock \href {https://doi.org/10.18653/v1/2020.acl-main.560} {The state and
  fate of linguistic diversity and inclusion in the {NLP} world}.
\newblock In \emph{Proceedings of the 58th Annual Meeting of the Association
  for Computational Linguistics}, pages 6282--6293, Online. Association for
  Computational Linguistics.

\bibitem[{Karthikeyan et~al.(2019)Karthikeyan, Wang, Mayhew, and
  Roth}]{karthikeyan2019cross}
K~Karthikeyan, Zihan Wang, Stephen Mayhew, and Dan Roth. 2019.
\newblock Cross-lingual ability of multilingual {BERT}: An empirical study.
\newblock In \emph{International Conference on Learning Representations}.

\bibitem[{Kingma and Ba(2015)}]{kingma2014adam}
Diederik~P. Kingma and Jimmy Ba. 2015.
\newblock \href {http://arxiv.org/abs/1412.6980} {Adam: {A} method for
  stochastic optimization}.
\newblock In \emph{3rd International Conference on Learning Representations,
  {ICLR} 2015, San Diego, CA, USA, May 7-9, 2015, Conference Track
  Proceedings}.

\bibitem[{Kondratyuk and Straka(2019)}]{kondratyuk-straka-2019-75}
Dan Kondratyuk and Milan Straka. 2019.
\newblock \href {https://doi.org/10.18653/v1/D19-1279} {75 languages, 1 model:
  Parsing universal dependencies universally}.
\newblock In \emph{Proceedings of the 2019 Conference on Empirical Methods in
  Natural Language Processing and the 9th International Joint Conference on
  Natural Language Processing (EMNLP-IJCNLP)}, pages 2779--2795, Hong Kong,
  China. Association for Computational Linguistics.

\bibitem[{Kudo and Richardson(2018)}]{kudo-richardson-2018-sentencepiece}
Taku Kudo and John Richardson. 2018.
\newblock \href {https://doi.org/10.18653/v1/D18-2012} {{S}entence{P}iece: A
  simple and language independent subword tokenizer and detokenizer for neural
  text processing}.
\newblock In \emph{Proceedings of the 2018 Conference on Empirical Methods in
  Natural Language Processing: System Demonstrations}, pages 66--71, Brussels,
  Belgium. Association for Computational Linguistics.

\bibitem[{Kundu et~al.(2018)Kundu, Paul, and Pal}]{kundu-etal-2018-deep}
Soumyadeep Kundu, Sayantan Paul, and Santanu Pal. 2018.
\newblock \href {https://doi.org/10.18653/v1/W18-2411} {A deep learning based
  approach to transliteration}.
\newblock In \emph{Proceedings of the Seventh Named Entities Workshop}, pages
  79--83, Melbourne, Australia. Association for Computational Linguistics.

\bibitem[{Kuratov and Arkhipov(2019)}]{kuratov-2019-rubert}
Yuri Kuratov and Mikhail Arkhipov. 2019.
\newblock \href {http://arxiv.org/abs/1905.07213} {Adaptation of deep
  bidirectional multilingual transformers for {R}ussian language}.
\newblock {arXiv}:1905.07213.

\bibitem[{Lin et~al.(2016)Lin, Pan, Deri, Ji, and
  Knight}]{lin-etal-2016-leveraging}
Ying Lin, Xiaoman Pan, Aliya Deri, Heng Ji, and Kevin Knight. 2016.
\newblock \href {https://www.aclweb.org/anthology/W16-2701} {Leveraging entity
  linking and related language projection to improve name transliteration}.
\newblock In \emph{Proceedings of the Sixth Named Entity Workshop}, pages
  1--10, Berlin, Germany. Association for Computational Linguistics.

\bibitem[{Liu et~al.(2019)Liu, Ott, Goyal, Du, Joshi, Chen, Levy, Lewis,
  Zettlemoyer, and Stoyanov}]{liu2019roberta}
Yinhan Liu, Myle Ott, Naman Goyal, Jingfei Du, Mandar Joshi, Danqi Chen, Omer
  Levy, Mike Lewis, Luke Zettlemoyer, and Veselin Stoyanov. 2019.
\newblock {RoBERTa}: A robustly optimized {BERT} pretraining approach.
\newblock {arXiv}:1907.11692.

\bibitem[{Liu et~al.(2020)Liu, Ott, Goyal, Du, Joshi, Chen, Levy, Lewis,
  Zettlemoyer, and Stoyanov}]{liu2020roberta}
Yinhan Liu, Myle Ott, Naman Goyal, Jingfei Du, Mandar Joshi, Danqi Chen, Omer
  Levy, Mike Lewis, Luke Zettlemoyer, and Veselin Stoyanov. 2020.
\newblock \href {https://openreview.net/forum?id=SyxS0T4tvS} {Ro{\{}bert{\}}a:
  A robustly optimized {\{}bert{\}} pretraining approach}.

\bibitem[{Martin et~al.(2020)Martin, Muller, Ortiz~Su{\'a}rez, Dupont, Romary,
  de~la Clergerie, Seddah, and Sagot}]{martin-etal-2020-camembert}
Louis Martin, Benjamin Muller, Pedro~Javier Ortiz~Su{\'a}rez, Yoann Dupont,
  Laurent Romary, {\'E}ric de~la Clergerie, Djam{\'e} Seddah, and Beno{\^\i}t
  Sagot. 2020.
\newblock \href {https://www.aclweb.org/anthology/2020.acl-main.645}
  {{C}amem{BERT}: a tasty {F}rench language model}.
\newblock In \emph{Proceedings of the 58th Annual Meeting of the Association
  for Computational Linguistics}, pages 7203--7219, Online. Association for
  Computational Linguistics.

\bibitem[{Micheli et~al.(2020)Micheli, d{'}Hoffschmidt, and
  Fleuret}]{Micheli_et_al:2020:smallbertmodels}
Vincent Micheli, Martin d{'}Hoffschmidt, and Fran{\c{c}}ois Fleuret. 2020.
\newblock \href {https://doi.org/10.18653/v1/2020.emnlp-main.632} {On the
  importance of pre-training data volume for compact language models}.
\newblock In \emph{Proceedings of the 2020 Conference on Empirical Methods in
  Natural Language Processing (EMNLP)}, pages 7853--7858, Online. Association
  for Computational Linguistics.

\bibitem[{Mikolov et~al.(2013)Mikolov, Sutskever, Chen, Corrado, and
  Dean}]{mikolov2013distributed}
Tomas Mikolov, Ilya Sutskever, Kai Chen, Greg~S Corrado, and Jeff Dean. 2013.
\newblock Distributed representations of words and phrases and their
  compositionality.
\newblock In \emph{Advances in neural information processing systems}, pages
  3111--3119.

\bibitem[{Murikinati et~al.(2020)Murikinati, Anastasopoulos, and
  Neubig}]{murikinati-etal-2020-transliteration}
Nikitha Murikinati, Antonios Anastasopoulos, and Graham Neubig. 2020.
\newblock \href {https://doi.org/10.18653/v1/2020.sigmorphon-1.22}
  {Transliteration for cross-lingual morphological inflection}.
\newblock In \emph{Proceedings of the 17th SIGMORPHON Workshop on Computational
  Research in Phonetics, Phonology, and Morphology}, pages 189--197, Online.
  Association for Computational Linguistics.

\bibitem[{Nivre et~al.(2016)Nivre, De~Marneffe, Ginter, Goldberg, Hajic,
  Manning, McDonald, Petrov, Pyysalo, Silveira et~al.}]{nivre2016universal}
Joakim Nivre, Marie-Catherine De~Marneffe, Filip Ginter, Yoav Goldberg, Jan
  Hajic, Christopher~D Manning, Ryan McDonald, Slav Petrov, Sampo Pyysalo,
  Natalia Silveira, et~al. 2016.
\newblock Universal dependencies v1: A multilingual treebank collection.
\newblock In \emph{Proceedings of the Tenth International Conference on
  Language Resources and Evaluation (LREC'16)}, pages 1659--1666.

\bibitem[{Ortiz~Su{\'a}rez et~al.(2020)Ortiz~Su{\'a}rez, Romary, and
  Sagot}]{suarez2020monolingual}
Pedro~Javier Ortiz~Su{\'a}rez, Laurent Romary, and Beno{\^\i}t Sagot. 2020.
\newblock \href {https://doi.org/10.18653/v1/2020.acl-main.156} {A monolingual
  approach to contextualized word embeddings for mid-resource languages}.
\newblock In \emph{Proceedings of the 58th Annual Meeting of the Association
  for Computational Linguistics}, pages 1703--1714, Online. Association for
  Computational Linguistics.

\bibitem[{{Ortiz Su{\'a}rez} et~al.(2019){Ortiz Su{\'a}rez}, Sagot, and
  Romary}]{OrtizSuarezSagotRomary2019}
Pedro~Javier {Ortiz Su{\'a}rez}, Beno{\^i}t Sagot, and Laurent Romary. 2019.
\newblock \href {https://doi.org/10.14618/ids-pub-9021} {Asynchronous pipelines
  for processing huge corpora on medium to low resource infrastructures}.
\newblock In \emph{Proceedings of the Workshop on Challenges in the Management
  of Large Corpora (CMLC-7) 2019. Cardiff, 22nd July 2019}, pages 9 -- 16,
  Mannheim. Leibniz-Institut f{\"u}r Deutsche Sprache.

\bibitem[{Pan et~al.(2017)Pan, Zhang, May, Nothman, Knight, and
  Ji}]{pan-etal-2017-cross}
Xiaoman Pan, Boliang Zhang, Jonathan May, Joel Nothman, Kevin Knight, and Heng
  Ji. 2017.
\newblock \href {https://doi.org/10.18653/v1/P17-1178} {Cross-lingual name
  tagging and linking for 282 languages}.
\newblock In \emph{Proceedings of the 55th Annual Meeting of the Association
  for Computational Linguistics (Volume 1: Long Papers)}, pages 1946--1958,
  Vancouver, Canada. Association for Computational Linguistics.

\bibitem[{Peters et~al.(2018)Peters, Neumann, Iyyer, Gardner, Clark, Lee, and
  Zettlemoyer}]{Peters:2018}
Matthew~E. Peters, Mark Neumann, Mohit Iyyer, Matt Gardner, Christopher Clark,
  Kenton Lee, and Luke Zettlemoyer. 2018.
\newblock Deep contextualized word representations.
\newblock In \emph{Proc. of NAACL}.

\bibitem[{Pfeiffer et~al.(2020)Pfeiffer, Vuli{\'c}, Gurevych, and
  Ruder}]{pfeiffer-etal-2020-mad}
Jonas Pfeiffer, Ivan Vuli{\'c}, Iryna Gurevych, and Sebastian Ruder. 2020.
\newblock \href {https://www.aclweb.org/anthology/2020.emnlp-main.617}
  {{MAD-X}: {A}n {A}dapter-{B}ased {F}ramework for {M}ulti-{T}ask
  {C}ross-{L}ingual {T}ransfer}.
\newblock In \emph{Proceedings of the 2020 Conference on Empirical Methods in
  Natural Language Processing (EMNLP)}, pages 7654--7673, Online. Association
  for Computational Linguistics.

\bibitem[{Pires et~al.(2019)Pires, Schlinger, and
  Garrette}]{pires2019multilingual}
Telmo Pires, Eva Schlinger, and Dan Garrette. 2019.
\newblock How multilingual is multilingual {BERT}?
\newblock In \emph{Proceedings of the 57th Annual Meeting of the Association
  for Computational Linguistics}, pages 4996--5001.

\bibitem[{Qi et~al.(2020)Qi, Zhang, Zhang, Bolton, and Manning}]{qi2020stanza}
Peng Qi, Yuhao Zhang, Yuhui Zhang, Jason Bolton, and Christopher~D. Manning.
  2020.
\newblock \href {https://doi.org/10.18653/v1/2020.acl-demos.14} {{S}tanza: A
  python natural language processing toolkit for many human languages}.
\newblock In \emph{Proceedings of the 58th Annual Meeting of the Association
  for Computational Linguistics: System Demonstrations}, pages 101--108,
  Online. Association for Computational Linguistics.

\bibitem[{Raffel et~al.(2019)Raffel, Shazeer, Roberts, Lee, Narang, Matena,
  Zhou, Li, and Liu}]{raffel2019exploring}
Colin Raffel, Noam Shazeer, Adam Roberts, Katherine Lee, Sharan Narang, Michael
  Matena, Yanqi Zhou, Wei Li, and Peter~J Liu. 2019.
\newblock Exploring the limits of transfer learning with a unified text-to-text
  transformer.
\newblock {arXiv}:1910.10683.

\bibitem[{Rahimi et~al.(2019)Rahimi, Li, and Cohn}]{rahimi-etal-2019-massively}
Afshin Rahimi, Yuan Li, and Trevor Cohn. 2019.
\newblock \href {https://doi.org/10.18653/v1/P19-1015} {Massively multilingual
  transfer for {NER}}.
\newblock In \emph{Proceedings of the 57th Annual Meeting of the Association
  for Computational Linguistics}, pages 151--164, Florence, Italy. Association
  for Computational Linguistics.

\bibitem[{Rijhwani et~al.(2019)Rijhwani, Xie, Neubig, and
  Carbonell}]{rijhwani2019zero}
Shruti Rijhwani, Jiateng Xie, Graham Neubig, and Jaime Carbonell. 2019.
\newblock Zero-shot neural transfer for cross-lingual entity linking.
\newblock In \emph{Proceedings of the AAAI Conference on Artificial
  Intelligence}, volume~33, pages 6924--6931.

\bibitem[{Schmidt and Wiegand(2017)}]{schmidt-wiegand-2017-survey}
Anna Schmidt and Michael Wiegand. 2017.
\newblock \href {https://doi.org/10.18653/v1/W17-1101} {A survey on hate speech
  detection using natural language processing}.
\newblock In \emph{Proceedings of the Fifth International Workshop on Natural
  Language Processing for Social Media}, pages 1--10, Valencia, Spain.
  Association for Computational Linguistics.

\bibitem[{Schweter(2020)}]{schweter-2020-berturk}
Stefan Schweter. 2020.
\newblock \href {https://doi.org/10.5281/zenodo.3770924} {{BERTurk} - {BERT}
  models for {T}urkish}.

\bibitem[{Seddah et~al.(2020)Seddah, Essaidi, Fethi, Futeral, Muller,
  Ortiz~Su{\'a}rez, Sagot, and Srivastava}]{seddah-etal-2020-building}
Djam{\'e} Seddah, Farah Essaidi, Amal Fethi, Matthieu Futeral, Benjamin Muller,
  Pedro~Javier Ortiz~Su{\'a}rez, Beno{\^\i}t Sagot, and Abhishek Srivastava.
  2020.
\newblock \href {https://doi.org/10.18653/v1/2020.acl-main.107} {Building a
  user-generated content {N}orth-{A}frican {A}rabizi treebank: Tackling hell}.
\newblock In \emph{Proceedings of the 58th Annual Meeting of the Association
  for Computational Linguistics}, pages 1139--1150, Online. Association for
  Computational Linguistics.

\bibitem[{Stecklow(2018)}]{myanmar2018}
Steve Stecklow. 2018.
\newblock {Why {F}acebook is losing the war on hate speech in {M}yanmar,
  {R}euters}.
\newblock
  \url{https://www.reuters.com/investigates/special-report/myanmar-facebook-hate/
  }.

\bibitem[{Straka(2018)}]{straka2018udpipe}
Milan Straka. 2018.
\newblock {UDP}ipe 2.0 prototype at {C}o{NLL} 2018 {UD} shared task.
\newblock In \emph{Proceedings of the CoNLL 2018 Shared Task: Multilingual
  Parsing from Raw Text to Universal Dependencies}, pages 197--207.

\bibitem[{Strubell et~al.(2019)Strubell, Ganesh, and
  McCallum}]{strubell-etal-2019-energy}
Emma Strubell, Ananya Ganesh, and Andrew McCallum. 2019.
\newblock \href {https://doi.org/10.18653/v1/P19-1355} {Energy and policy
  considerations for deep learning in {NLP}}.
\newblock In \emph{Proceedings of the 57th Annual Meeting of the Association
  for Computational Linguistics}, pages 3645--3650, Florence, Italy.
  Association for Computational Linguistics.

\bibitem[{de~Vries et~al.(2019)de~Vries, van Cranenburgh, Bisazza, Caselli, van
  Noord, and Nissim}]{de2019bertje}
Wietse de~Vries, Andreas van Cranenburgh, Arianna Bisazza, Tommaso Caselli,
  Gertjan van Noord, and Malvina Nissim. 2019.
\newblock Bertje: A dutch {BERT} model.
\newblock {arXiv}:1912.09582.

\bibitem[{Wolf et~al.(2020)Wolf, Debut, Sanh, Chaumond, Delangue, Moi, Cistac,
  Rault, Louf, Funtowicz, Davison, Shleifer, von Platen, Ma, Jernite, Plu, Xu,
  Le~Scao, Gugger, Drame, Lhoest, and Rush}]{wolf-etal-2020-transformers}
Thomas Wolf, Lysandre Debut, Victor Sanh, Julien Chaumond, Clement Delangue,
  Anthony Moi, Pierric Cistac, Tim Rault, Remi Louf, Morgan Funtowicz, Joe
  Davison, Sam Shleifer, Patrick von Platen, Clara Ma, Yacine Jernite, Julien
  Plu, Canwen Xu, Teven Le~Scao, Sylvain Gugger, Mariama Drame, Quentin Lhoest,
  and Alexander Rush. 2020.
\newblock \href {https://www.aclweb.org/anthology/2020.emnlp-demos.6}
  {Transformers: State-of-the-art natural language processing}.
\newblock In \emph{Proceedings of the 2020 Conference on Empirical Methods in
  Natural Language Processing: System Demonstrations}, pages 38--45, Online.
  Association for Computational Linguistics.

\bibitem[{Wu and Dredze(2020)}]{wu-dredze-2020-languages}
Shijie Wu and Mark Dredze. 2020.
\newblock \href {https://www.aclweb.org/anthology/2020.repl4nlp-1.16} {Are all
  languages created equal in multilingual {BERT}?}
\newblock In \emph{Proceedings of the 5th Workshop on Representation Learning
  for NLP}, pages 120--130, Online. Association for Computational Linguistics.

\bibitem[{Zeman and Haji{\v{c}}(2018)}]{conll-2018-conll}
Daniel Zeman and Jan Haji{\v{c}}, editors. 2018.
\newblock \href {https://www.aclweb.org/anthology/K18-2000} {\emph{Proceedings
  of the {C}o{NLL} 2018 Shared Task: Multilingual Parsing from Raw Text to
  {U}niversal {D}ependencies}}. Association for Computational Linguistics,
  Brussels, Belgium.

\end{thebibliography}
\bibliographystyle{acl_natbib}


\appendix

\section{Languages}

We list the 15 typologically diverse unseen languages we experiment with in Table~\ref{tab:langs}  with information on their language family, script, origin and number of sentences available along with the categories we classified them in.

\paragraph{Data Sources}

We base our experiments on data originated from two sources. The Universal Dependency project \cite{nivre2016universal} downloadable here \url{https://lindat.mff.cuni.cz/repository/xmlui/handle/11234/1-2988} and the WikiNER dataset \cite{pan-etal-2017-cross}. We also \draftremove{make} use of the CoNLL-2003 shared task NER English dataset \url{https://www.clips.uantwerpen.be/conll2003/}

\section{Reproducibility}
\label{sec:reproducibility}

\paragraph{Infrastructure}

Our experiments were ran on a shared cluster on the equivalent of 15 Nvidia Tesla T4 GPUs.\footnote{https://www.nvidia.com/en-sg/data-center/tesla-t4/}

\paragraph{Optimization}

\begin{table}[h!]
\footnotesize 
\centering
\begin{tabular}{lrrrr}
\toprule
  \textit{Params.}& Parsing  &  NER & POS & Bounds\\
  \hline
  batch size & 32 & 16 & 16 & [1,256] \\
  learning rate &  5e-5 & 3.5e-5 & 5e-5 & [1e-6,1e-3] \\
  epochs (best) &  15 & 6 & 6 & [1,$+\inf$] \\
  \#grid & 60 & 60 & 180 & - \\
  Run-time & 32 & 24 & 75 & -\\
\bottomrule

\end{tabular}
\caption{Fine-tuning best hyper-parameters for each task as selected on the validation set with bounds. \#grid: number of grid search trial. Run-time is reported in average for training and evaluation. Run-time indicated in minutes.}
 
\label{tab:hyperparameters}
\end{table}
\begin{table}[h!]
\footnotesize
\centering
\scalebox{1}{
\begin{tabular}{lrr}
\toprule
   Parameter  & Value \\
  \midrule
  batch size & 64 \\
  learning rate &  5e-5  \\
  optimizer & Adam \\
  warmup & linear \\
  warmup steps & 10\% total \\
  epochs (best of)  & 10 \\
\bottomrule
\end{tabular}
}
\caption{Unsupervised fine-tuning hyper-parameters}
\label{tab:unsupervisedhyperparamters}
\end{table}
For all pretraining and fine-tuning runs, we use the Adam optimizer  \citep{kingma2014adam}. 
For fine-tuning, following \citet{devlin-etal-2019-bert}, we only back-propagate through the first token of each word. We select the hyperparameters that \draftreplace{minmize}{minimize} the loss on the validation set. The reported results are the average score of 5 runs with different random seeds computed on the test splits. 
We report the hyperparameters in Table~\ref{tab:hyperparameters}-\ref{tab:unsupervisedhyperparamters}\draftremove{  for \tasktuning in the Appendix in Table~\ref{tab:hyperparameters} and about pretraining and \mlmtuning in Table~\ref{tab:unsupervisedhyperparamters}}.

\section{Easy Languages}
\label{sec:easy_appendix}
We describe here in more details languages that we classify as Easy 
in section~\ref{sec:easy}.

In practice, one can obtain very high performance even in zero-shot settings for such languages, by performing task-tuning on related languages. 

\begin{table}[htbp]
    \centering\small
    \begin{tabular}{clccc}
    \toprule
        & Model & UPOS & LAS & NER \\
    \midrule
    \multicolumn{5}{l}{\textit{Zero-Shot}}\\
        (1) &FaroeseBERT & 66.4 & 35.8 & - \\
        (2) & \mbert & 79.4 & 67.5 & -\\
        (3) & \mbert+MLM & \textbf{83.4} & \textbf{67.8} &-  \\
    \midrule
    \multicolumn{5}{l}{\textit{Few-Shot (CV with around 500 instances)}}\\
    (4) & Baseline & 95.36 & 83.02 & 44.8 \\
    (5) & FaroeseBERT &91.12 & 67.66 & 39.3 \\
    (6) & \mbert & 96.31 &  84.02 & 52.1 \\
    (7) & \mbert+MLM &\textbf{96.52} & \textbf{86.41} & \textbf{58.3} \\
    
    \bottomrule
    \end{tabular}
    \caption{Faroese is an ``easy'' unseen language: a multilingual model (+ language-specific MLM) easily outperforms all baselines. Zero-shot performance, after task-tuning only on related languages (Danish, Norwegian, Swedish) is also high.}
    \label{tab:faroese}
\end{table}

Perhaps the best example of such an ``easy'' setting is Faroese. mBERT has been trained on several languages of the north Germanic genus of the Indo-European language family, all of which use the Latin script. As a result, the multilingual \mbert model performs much better than the monolingual FaroeseBERT model that we trained on the available Faroese text (cf rows 1--2 and 5--6 in Table~\ref{tab:faroese}). Fine-tuning \mbert on the Faroese text is even more effective (rows 3 and 6 in Table~\ref{tab:faroese}), leading to further improvements, reaching more than 96.5\% POS-tagging accuracy, 86\% LAS for dependency parsing, and 58\% NER F1 in the few-shot setting, surpassing the non-contextual baseline. In fact, even in zero-shot conditions, where we task-tune only on related languages (Danish, Norwegian, and Swedish), the model achieves remarkable performance of over 83\% POS-tagging accuracy and 67.8\% LAS dependency parsing.

\begin{table}[htbp]
    \centering\small
    \begin{tabular}{clcc}
    \toprule
        & Model & UPOS & LAS\\
    \midrule
    \multicolumn{4}{l}{\textit{Zero-Shot}}\\
        (1) & SwissGermanBERT & 64.7 & 30.0 \\
        (2) & \mbert & 62.7 & 41.2\\
        (3) & \mbert+MLM & \textbf{87.9} & \textbf{69.6}\\
    \midrule
    \multicolumn{4}{l}{\textit{Few-Shot (CV with around 100 instances)}}\\
        (4) & Baseline & 75.22 & 32.18 \\
        (5) & SwissGermanBERT & 65.42 & 30.0 \\
        (6) & \mbert & 76.66 & 41.2\\
        (7) & \mbert+MLM & \textbf{78.68} & \textbf{69.6}\\
        
    \bottomrule
    \end{tabular}
    \caption{Swiss German is an ``easy'' unseen language: a multilingual model (+ language-specific MLM) outperforms all baselines in both zero-shot (task-tuning on the related High German) and few-shot settings.}
    \label{tab:swiss}
\end{table}
Swiss German is another example of a language for which one can easily adapt a multilingual model and obtain good performance even in zero-shot settings. As in Faroese, simple MLM fine-tuning of the \mbert model with 200K sentences leads to an improvement of more than 25 points in both POS tagging and dependency parsing (Table~\ref{tab:swiss}) in zero-shot settings, with similar improvement trends in the few-shot setting. 

The potential of similar-language pretraining along with script similarity is also showcased in the case of Naija (also known as Nigerian English or Nigerian Pidgin), an English creole spoken by millions in Nigeria. As Table~\ref{tab:naija} shows, with results after language- and task-tuning on 6K training examples, the multilingual approach surpasses the monolingual baseline.

\begin{table}[htbp]
    \centering\small
    \begin{tabular}{lcc}
    \toprule
        Model & UPOS & LAS\\
    \midrule
        NaijaBERT & 87.1 & 63.02 \\
        \mbert & 89.3 & \textbf{71.6}\\
        \mbert+MLM & \textbf{89.6} & 69.2\\
    \bottomrule
    \end{tabular}
    \caption{Performance on Naija, an English creole, is very high, so we also classify it as an ``easy'' unseen language.}
    \label{tab:naija}
\end{table}

On a side note, we can rely on ~\citet{han-eisenstein-2019-unsupervised} to also classify Early Modern English as an easy language. Similarly, the work of~\citet{chau2020parsing} allows us to  also classify Singlish (Singaporean English) as an easy language. In both cases, these languages are technically unseen by mBERT, but the fact that they are variants of English allows them to be easily handled by \mbert.

\section{Additional Uralic languages experiments}
\label{sec:uralic}

Following a similar procedure as in the Appendix~\ref{sec:easy_appendix}, we start with \mbert, perform task-tuning on Finnish and Estonian (both of which use the Latin script) and then do zero-shot experiments on Livvi, and Komi, all low-resource Uralic languages (results on the top part of Table~\ref{tab:uralic}). We also report results on the Finnish treebanks after task-tuning, for better comparison. The difference in performance on Livvi (which uses the Latin script) and the other languages that use the Cyrillic script is striking.

\begin{table}[ht!]
    \centering\small
    \begin{tabular}{clcc}
    \toprule
        & Language & UPOS & LAS\\
    \midrule
    \multicolumn{4}{l}{\textit{Task-tuned -- Latin script}}\\
        &Finnish (FTB) & 93.1 & 77.5\\
        &Finnish (TDT) & 95.0 & 78.9\\
        &Finnish (PUD) & 96.8 & 83.5\\
    \midrule
    \multicolumn{4}{c}{Zero-Shot Experiments}\\
    \multicolumn{4}{l}{\textit{Latin script}}\\
        &Livvi & 72.3 & 40.3\\
    \multicolumn{4}{l}{\textit{Cyrillic script}}\\
        &Erzya & 51.5 & 18.6\\
    \midrule
    \multicolumn{4}{c}{Few-Shot Experiments (CV)}\\
        \multicolumn{4}{l}{Livvi -- \textit{Latin script}}\\
         &Baseline & 84.1 & 40.1 \\
          &\mbert & 83.0 & 36.3 \\
          &\mbert+MLM & \textbf{85.5} & \textbf{42.3} \\
    \multicolumn{4}{l}{Erzya -- \textit{Cyrillic script}}\\
         &Baseline & 91.1 & 65.1 \\
         &\mbert & 89.3 & 61.2 \\
         &\mbert+MLM & \textbf{91.2} & \textbf{66.6} \\
    \bottomrule
    \end{tabular}
    \caption{The script matters for the efficacy of cross-lingual transfer. The zero-shot performance on Livvi, which is written in the same script as the task-tuning languages (Finnish, Estonian), is almost twice as good as the performance on the Uralic languages that use the Cyrillic script.}
    \label{tab:uralic}
\end{table}

Although they are not easy enough to be tackled in a zero-shot setting, we show that the low-resource Uralic languages fall in the  ``Intermediate'' category, since \mbert has been trained on similar languages: a small amount of annotated data are enough to improve over \mbert using task-tuning.

For both Livvi and Erzya, the multilingual model along with \mlmtuning achieves the best performance, outperforming the non-contextual baseline by more than 1.5 point for parsing and POS tagging.

\section{Controlled experiment: Transliterating  High-Resource Languages}
\label{sec:controlled}

To have a broader view on the effect of transliteration when using \mbert (section~\ref{sec:translit}), we study the impact of transliteration to the Latin script on high resource languages seen during \mbert pretraining such as Arabic, Japanese and Russian. We compare the performance of \mbert fine-tuned and evaluated on the original script with \mbert fine-tuned and evaluated on the transliterated text. As reported in Table~\ref{tab:controlled_exp}, transliterating those languages to the Latin script leads to large drop in performance for all the three tasks. 

\begin{table*}[!]
\footnotesize\centering
\scalebox{0.83}{ 
\begin{tabular}{lrrrrr}
    \toprule
    
    Language (iso) & Script & Family & \#sents & source & Category\\
    \midrule

    Faroese (fao) & Latin & North Germanic  &297K & \citep{biemann2007leipzig} & Easy \\ 
    Mingrelian (xmf) & Georg. & Kartvelian & 29K & Wikipedia  & Easy\\
    Naija (pcm) & Latin & English Pidgin & 237K & \citep{caron2019surface} & Easy\\
    Swiss German (gsw) & Latin & West Germanic & 250K & OSCAR  & Easy\\
    
    Bambara (bm) & Latin & Niger-Congo& 1K & OSCAR  & Intermediate\\
    Wolof  (wo) & Latin & Niger-Congo & 10K & OSCAR  & Intermediate\\
    Narabizi (nrz) & Latin & Semitic*
    & 87K &  \citep{seddah-etal-2020-building} & Intermediate\\
    Maltese (mlt) & Latin & Semitic & 50K  & OSCAR& Intermediate\\
    Buryat (bxu) & Cyrillic & Mongolic & 7K & Wikipedia & Intermediate\\
    Mari (mhr) & Cyrillic & Uralic & 58K & Wikipedia & Intermediate\\
    Erzya (myv) & Cyrillic & Uralic &  20K& Wikipedia & Intermediate\\
    Livvi (olo) & Latin & Uralic &  9.4K& Wikipedia & Intermediate\\
    
    Uyghur (ug) & Arabic & Turkic &  105K & OSCAR & Hard\\
    Sindhi (sd) & Arabic & Indo-Aryan & 375K & OSCAR & Hard\\
        Sorani (ckb) & Arabic & Indo-Iranian
    & 380K & OSCAR & Hard\\


\bottomrule
\end{tabular}
}
\caption{Unseen Languages used for our experiments. \#sents indicates the number of sentences used for training from scratch Monolingual Language Models as well as for \mlmtuning  \mbert  
\\ 
\small{*code-mixed with French}}
\label{tab:langs}
\end{table*}

\end{document}